\PassOptionsToPackage{dvipsnames}{xcolor} 
\PassOptionsToPackage{vlined,ruled,linesnumbered}{algorithm2e} 
\documentclass{article}

\usepackage{amsmath,amsfonts,bm}

\def\eqref#1{equation~\ref{#1}}

\def\1{\bm{1}}
\newcommand{\train}{\mathcal{D}}

\newcommand{\G}{\boldsymbol{G}}
\newcommand{\C}{\boldsymbol{C}}
\newcommand{\nn}{\boldsymbol{n}}
\newcommand{\X}{\boldsymbol{X}}

\newcommand{\Y}{\boldsymbol{Y}}
\newcommand{\W}{\boldsymbol{W}}
\newcommand{\K}{\boldsymbol{\mathcal{S}}}
\newcommand{\T}{\top}

\DeclareMathAlphabet{\mathsfit}{\encodingdefault}{\sfdefault}{m}{sl}
\SetMathAlphabet{\mathsfit}{bold}{\encodingdefault}{\sfdefault}{bx}{n}

\newcommand{\R}{\mathbb{R}}

\usepackage{microtype}
\usepackage{graphicx}
\usepackage{subfigure}
\usepackage{booktabs} 
\usepackage{graphicx,adjustbox}
\usepackage{url}
\usepackage{amsthm}
\usepackage[linesnumbered,ruled,vlined]{algorithm2e}
\SetKwComment{Comment}{/* }{ */}
\SetAlFnt{\small}
\SetKwBlock{Initialization}{Initialization:}{}
\RestyleAlgo{ruled}
\usepackage{amsmath}
\usepackage{array}
\usepackage{multirow}
\usepackage{booktabs}
\usepackage[table]{xcolor}
\usepackage[normalem]{ulem}
\useunder{\uline}{\ul}{}
\usepackage{float, wrapfig}
\usepackage{resizegather}
\usepackage{enumitem}
\usepackage{color,xcolor}
\definecolor{LightCyan}{rgb}{0.88,1,1}
\usepackage{hyperref}

\newcommand{\methodshort}[1]{STSA}
\newcommand{\methodshortvar}[1]{STSA-E}
\newcommand{\methodall}[1]{STSA}

\usepackage[accepted]{icml2025}

\usepackage{amsmath}
\usepackage{amssymb}
\usepackage{mathtools}
\usepackage{amsthm}

\usepackage[capitalize,noabbrev]{cleveref}

\theoremstyle{plain}

\theoremstyle{definition}

\theoremstyle{remark}

\usepackage[textsize=tiny]{todonotes}

\icmltitlerunning{Enhancing Federated Class-Incremental Learning via
Spatial-Temporal Statistics Aggregation}

\begin{document}

\twocolumn[
    \icmltitle{Enhancing Federated Class-Incremental Learning via \\ Spatial-Temporal
Statistics Aggregation}


    \icmlsetsymbol{equal}{*}

    \begin{icmlauthorlist}
        \icmlauthor{Zenghao Guan}{ucas,key}
        \icmlauthor{Guojun Zhu}{ucas}
        \icmlauthor{Yucan Zhou}{tju}
        \icmlauthor{Wu Liu}{ustc}
        \icmlauthor{Weiping Wang}{key}
        \icmlauthor{Jiebo Luo}{rocherster}
        \icmlauthor{Xiaoyan Gu}{ucas,key}
    \end{icmlauthorlist}

    \icmlaffiliation{ucas}{University of Chinese Academy of Sciences}
    \icmlaffiliation{tju}{Tianjin University}
    \icmlaffiliation{ustc}{University of Science and Technology of China}
    \icmlaffiliation{rocherster}{University of Rochester}
    \icmlaffiliation{key}{State Key Laboratory of Cyberspace Security Defense, Institute of Information Engineering}
    
    \icmlkeywords{Machine Learning}

    \vskip 0.3in
]



\printAffiliationsAndNotice{}  

\begin{abstract}
Federated Class-Incremental Learning (FCIL) enables Class-Incremental Learning (CIL) from distributed data. Existing FCIL methods typically integrate old knowledge preservation into local client training. However, these methods cannot avoid spatial-temporal client drift caused by data heterogeneity and often incur significant computational and communication overhead, limiting practical deployment. To address these challenges simultaneously, we propose a novel approach, \textbf{S}patial-\textbf{T}emporal \textbf{S}tatistics \textbf{A}ggregation (\methodshort{}), which provides a unified framework to aggregate feature statistics both spatially (across clients) and temporally (across stages). The aggregated feature statistics are unaffected by data heterogeneity and can be used to update the classifier in closed form at each stage. Additionally, we introduce \methodshortvar{}, a communication-efficient variant with theoretical guarantee, achieving similar performance to \methodshort{} with much lower communication overhead. Extensive experiments on three widely used FCIL datasets, with varying degrees of data heterogeneity, show that our method outperforms state-of-the-art FCIL methods in terms of performance, flexibility, and both communication and computation efficiency. The code is available at \url{https://github.com/Yuqin-G/STSA}.

\end{abstract}

\section{Introduction}
Federated Learning (FL) \cite{fedavg} is an emerging distributed machine learning framework that enables multiple parties to participate in collaborative learning \cite{liqi, liu2024lmagent,liu2025popsim,liu2025rumorsphere, li2026drgwlearningdisentangledrepresentations} under the coordination of a central server. The central server aggregates the model updates uploaded by clients rather than the private data. This mechanism addresses the data silos problem and provides privacy benefits for distributed learning \cite{fls1}. 

Existing FL methods \cite{fedavg, fcl} typically assume that data categories and domains remain static throughout the entire training and testing phases. However, this assumption fails to align with real-world environments, where data distributions are dynamic and continuously evolving \cite{pfcil_prompt, fdl1, fcl_survey, fcl_survey2}. To tackle FL tasks in dynamic environments, recent research has incorporated Continual Learning (CL) techniques into the FL framework, giving rise to the approach known as Federated Continual Learning (FCL) \cite{fcl, fclkd, fclwt}. In FCL, Federated Class-Incremental Learning (FCIL) \cite{target,pilora,lander} has gained significant attention in recent years due to its wide applicability \cite{target, privacy1}. This setting allows new classes to be added flexibly at any time. In this paper, we focus on the global FCIL setting \cite{fcl, fcl_lga, lander, target, pilora}, where a single global model is collaboratively trained across clients to learn new classes over time, aiming to enhance the generalization of the global model across all encountered classes.

\begin{figure}[t]
 \centering
 \vspace{-4pt}
\includegraphics[width=1.0\linewidth]{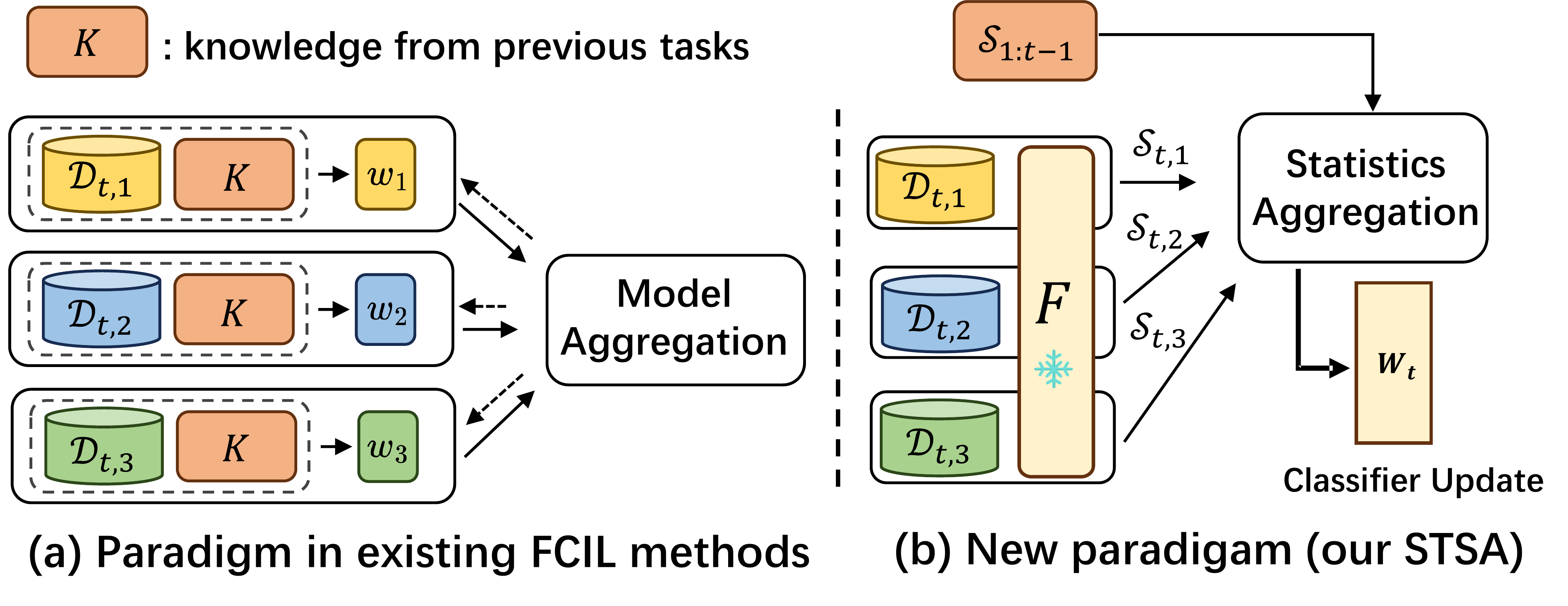}
\vspace{-15pt}
\caption{Paradigms of existing FCIL methods and ours. $w$ and $\mathcal{S}$ denote the model weights and feature statistics, respectively.}
\label{fig:motivation1}
\end{figure}

\begin{figure}[t]
 \centering
 \vspace{-8pt}
\includegraphics[width=1.0\linewidth]{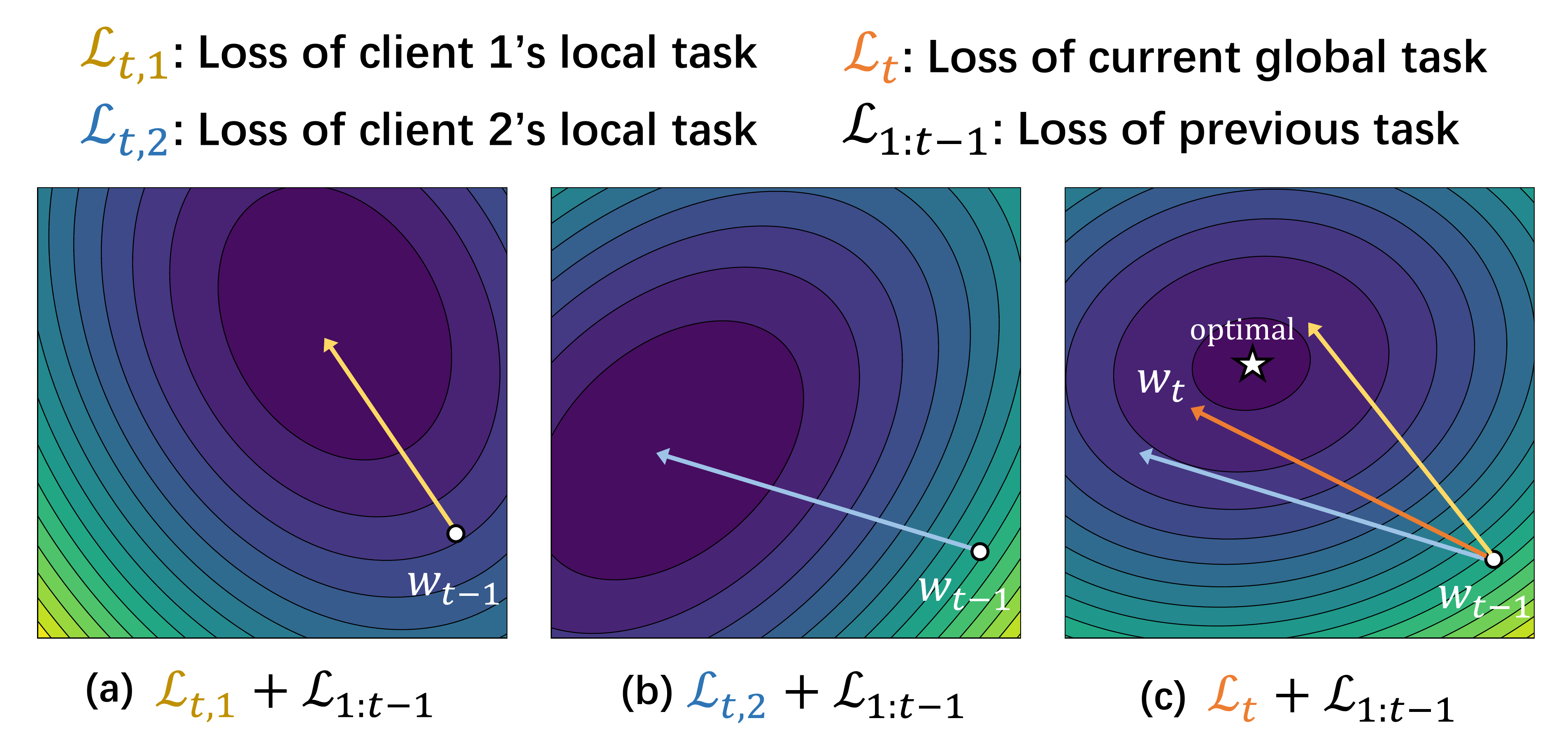}
\vspace{-15pt}
\caption{Illustration of spatial-temporal client drift through loss landscapes. $W_t$ denotes the aggregated global model.}
\label{fig:drift}
\end{figure}

In FCIL, we need to simultaneously address two core challenges: catastrophic forgetting and data heterogeneity. Existing FCIL methods typically follow the paradigm where the retention of old knowledge is integrated directly into local client training through techniques such as experience replay \cite{fcl, fcl_lga, fcl_replay1, target, fedcil, mfcl, lander}, weight regularization \cite{fl_clip, fedssi}, and parameter-efficient fine-tuning \cite{fcl_prompt, pilora, fcl_prompt2} (as shown in Figure \ref{fig:motivation1}). However, this paradigm still suffers from two critical drawbacks that limit its practical deployment. \textcircled{1} \textbf{Spatial-temporal client drift} caused by data heterogeneity. Specifically, each client has a distinct data distribution, leading to different balances in the optimization process between learning new tasks and preserving old knowledge. This disparity causes inconsistent local updates, degrading the aggregated global model's generalizability to both new and old tasks. As shown in Figure \ref{fig:drift}, we present the loss landscapes of the local and global models. The local model performs well on old tasks and current local tasks, but the global model shows a noticeable drift with the optimal model. \textcircled{2} \textbf{High communication and computation overhead}. Existing methods often necessitate iterative local updates and numerous communication rounds between clients and the server. This imposes significant computational burdens on client devices and leads to high communication overhead, posing practical limitations, especially for resource-constrained clients. These challenges motivated us to explore: \textit{whether it is possible to break this paradigm and design a new method that effectively mitigates spatial-temporal client drift caused by data heterogeneity while maintaining low computational and communication overhead.}

In this paper, we propose a method named \textbf{S}patial-\textbf{T}emporal \textbf{S}tatistics \textbf{A}ggregation (\methodshort{}). Instead of using aggregated local updates to update the global model, \methodshort{} aggregates feature statistics spatially (across clients) to learn new tasks and temporally (across stages) to retain old knowledge, using a fixed global feature extractor. Specifically, in the first stage of \methodshort{}, all clients engage in federated training to obtain a global model. The trained backbone is then fixed and distributed among all clients for feature statistics extraction at each subsequent stage. These statistics are first aggregated across clients and then across stages. Notably, the aggregated feature statistics are equivalent to those derived from all data samples across both dimensions due to their linear characteristics, so \textbf{they are not affected by data heterogeneity.} Finally, these aggregated statistics are used to update the classifier in closed form at each stage. Since our method \textbf{requires no training} after the first stage and only needs one round of communication to upload feature statistics, it significantly improves computation and communication efficiency. Moreover, we propose a more communication-efficient variant, termed \methodshortvar{}, which enables the server to approximate global second-order feature statistics based on first-order statistics transmitted from clients. Theoretical analysis shows that \methodshortvar{} achieves similar statistical performance to \methodshort{} while considerably reducing communication overhead. 

Extensive experiments on three widely used FCIL datasets demonstrate the superiority of our methods. In terms of performance, our \methodshort{} achieves over \textbf{10\%} absolute improvements compared to the best-performing parameter-efficient fine-tuning baseline. Moreover, \methodshort{} achieves remarkable efficiency improvements, drastically reducing both communication and computation overhead. In particular, it delivers nearly \textbf{100×} faster computation compared to existing methods. 

In summary, the main contributions of this paper are three-fold:

\begin{itemize}
\item We propose a novel FCIL method called STSA, which offers a unified way to aggregate feature statistics both spatially (across clients) and temporally (across stages). The aggregated feature statistics are immune to data heterogeneity and can be used to update the classifier in closed form at each stage.
\item 
We propose STSA-E, a communication-efficient variant where the server approximates global second-order feature statistics using first-order statistics sent by clients. Theoretical and experimental results show that it achieves comparable performance to \methodshort{} while significantly reducing communication overhead.
\item We conduct extensive experiments to validate our method. The results demonstrate significant improvements in performance, flexibility, and both communication and computational efficiency compared to state-of-the-art FCIL methods.
\vspace{-10pt}
\end{itemize}

\begin{figure*}[t]
 \centering
 \vspace{-8pt}
\includegraphics[width=1.0\linewidth]{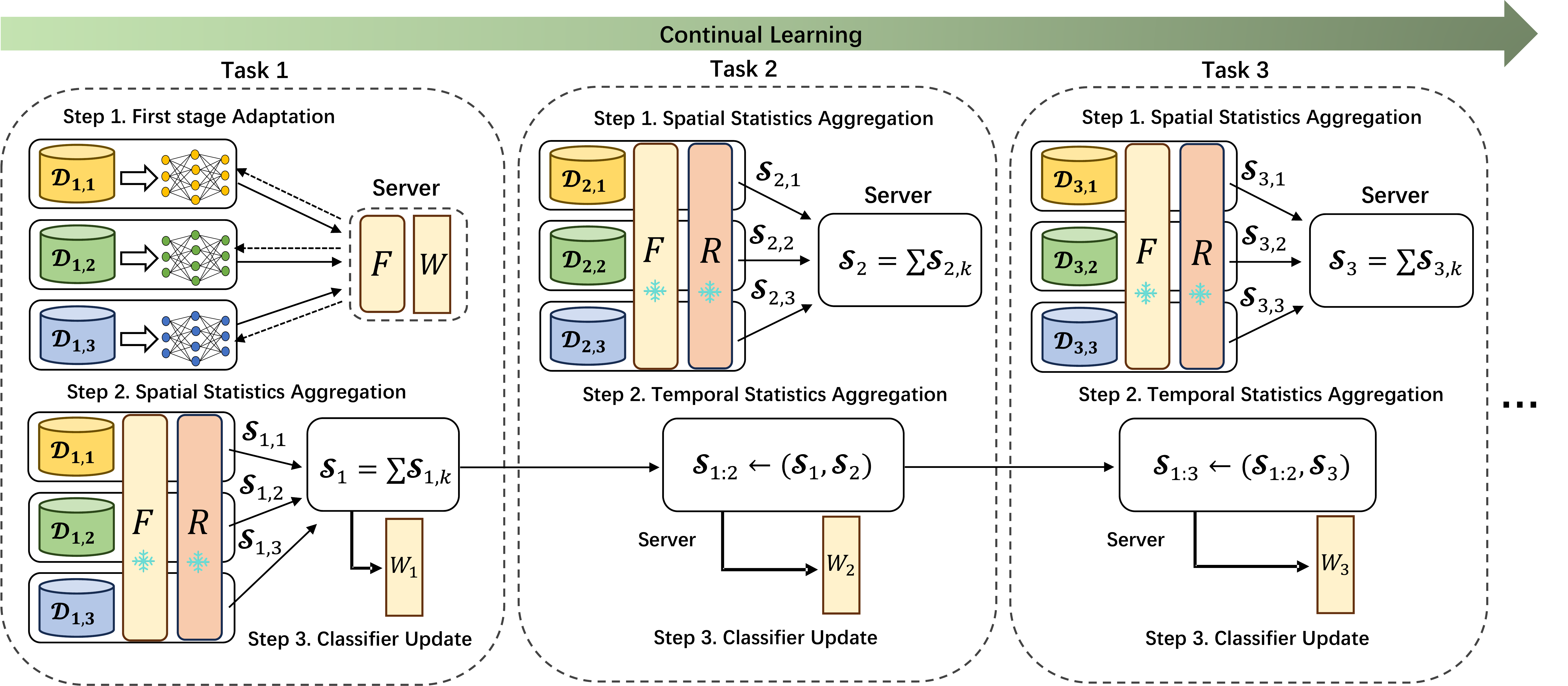}
\vspace{-15pt}
\caption{Framework of our algorithm. $\mathcal{D}_{i,j}$ denotes the local dataset of client $j$ for task $\mathcal{T}_i$. $F$ is the feature extractor trained in the first stage and is kept fixed afterward. $R$ is the random mapping layer. In our \methodshort{}, Aggregated Spatial Statistics $\K_{t}$ at stage $t$ includes both $\G_{t}$ and $\C_{t}$, whereas for \methodshortvar{}, it refers to $\C_{t}$.}
\label{fig:framework}
\vspace{-10pt}
\vspace{-5pt}
\end{figure*}

\section{Method}
\subsection{Problem Definition}
In FCIL, we have a set of $T$ tasks ~$\mathcal{T}=\{\mathcal{T}_t\}_{t=1}^T$ shared among $K$ clients, where the set of classes $\mathcal{C}_t$ ($|\mathcal{C}_t|=c_t$) available at each time step $t$ is disjoint. Each client $k$ only have access to private dataset $\small \train_{t,k}$ 
with size $n_{t,k}$ at task $\mathcal{T}_t$. Here we denote $\nn_{t,k}\in\R^{c_t}$ as the label frequency. During the training of task $\mathcal{T}_t$, each local client $k$ trains its local model $\theta_{t,k}$ using its local data $\train_{t,k}$ and then uploads the updated model parameters to the server. The server aggregates the local models to obtain the global model $\theta_t$ and distribute it to all clients as the initialization for the next round of training. This process is repeated for multiple rounds, and the final global model $\theta_t$ can distinguish the samples belonged to the classes set $\bigcup_{i=1}^{t}\mathcal{C}_i$. 

\subsection{Spatial-Temporal Statistics Aggregation (STSA)}

\noindent
\textbf{First Stage Adaptation.}
At task $\mathcal{T}_1$, each client participates in federated training to obtain a global model $\theta_1$. Any federated learning algorithm can be used in this phase. After training, we freeze the feature extractor $F$ from the global model $\theta_1$ for further use.

\vspace{-3pt}
\noindent
\textbf{Spatial Statistics Aggregation.}
For each task $\mathcal{T}_t$ ($t>0$), each client $k$ uses this fixed feature extractor $F$ to extract features from its local dataset. Given that the extracted features may lack sufficient expressiveness for clear class separation \cite{cover2006geometrical, ranpac}, we apply a nonlinear random mapping to project the extracted features into a higher-dimensional space (from $d$ to $M$, $M > d$):
\begin{align}
\X_{t,k} = & \begin{bmatrix}
\Phi(x_1),\Phi(x_2),\cdots,\Phi(x_{n_{t,k}})
\end{bmatrix}^\T,
\end{align}
where $\{(x_i, y_i)\}_{i=1}^{n_{t,k}} = \train_{t,k}$, $\Phi = R \circ F$ represents the composition of the feature extractor $F$ and the random mapping layer $R$, and $\X_{t,k}\in \mathbb{R}^{n_{t,k}\times M}$ denotes the concatenation of all feature vectors. Here, $R$ consists of a random matrix followed by a ReLU function. This $R$ is fixed after initialization and shared across all clients. To reduce storage overhead in the clients, the server can distribute a common \texttt{seed} to all clients. This allows clients to generate the same random matrix locally using the \texttt{seed}. After obtaining high-dimensional features, clients can delete it and regenerate it when needed.

Subsequently, each client computes the local gram matrix $\G_{t,k}\!=\!\!\X_{t, k}^\T\X_{t, k}\in \mathbb{R}^{M\times M}$ and the feature-label correlation matrix $\C_{t,k}\!=\!\!\X_{t, k}^\T \Y_{t, k}\in \mathbb{R}^{M\times c_t}$. Here $\Y_{t,k}$ denotes the corresponding label matrix stacked by one-hot vectors. The Spatial Statistics $\K_{t,k}=\{\G_{t,k},\C_{t,k}\}$ of each client $k$ is then uploaded to the server and aggregated as follows:
\begin{align}
\label{spatial_aggregation}
\G_t=\sum_{k=1}^K \G_{t,k}, \quad\C_t=\sum_{k=1}^K \C_{t,k}.
\end{align}
Due to the linear additivity of these statistics, we have:
\begin{align}    \G_{\text{centralized}}=\!\!\!\!\sum_{(x,y)\in\train_t}&\!\!\!\!\Phi(x)^\T\Phi(x) \notag\\
=\sum_{k=1}^K\;\sum_{(x,y)\in\train_{t,k}}&\!\!\!\Phi(x)^\T\Phi(x)=\sum_{k=1}^K\G_{t,k}=\G_t. \notag \\
\C_{\text{centralized}}=\!\!\!\!\sum_{(x,y)\in\train_t}&\!\!\!\!\Phi(x)^\T\text{OneHot}(y)  \notag\\
=\sum_{k=1}^K\;\sum_{(x,y)\in\train_{t,k}}\!\!\!\Phi(x)&^\T\text{OneHot}(y)=\sum_{k=1}^K\C_{t,k}=\C_t
\end{align}
This demonstrates that the statistics derived from Spatial Statistics Aggregation are consistently equivalent to those from the centralized setting (with access to all data from all clients at stage $t$). Therefore, these statistics are invariant to varing levels of data heterogeneity.

\noindent
\textbf{Temporal Statistics Aggregation.}
The server then aggregates the Aggregated Spatial Statistics $\K_{t}=\{\G_t,\C_t\}$ with Temporal Statistics $\K_{1:t-1}=\{\G_{1:t-1},\C_{1:t-1}\}$, respectively:
\begin{align}
\label{temporal_aggregation}
    \G_{1:t}=\G_{1:t-1}+\G_{t},\quad \C_{1:t}=[\C_{1:t-1},\C_{t}],
\end{align}
where $\G_{1:t-1}\!\!=\!\!\sum_{i=1}^{t-1}\G_i$ and $\C_{1:t-1}\!=\![\C_1,\C_2,\!\cdots,\!\C_{t-1}]$ are accumulated along $1\sim(t-1)$ incremental stages. $\G_{1:t}$ and $\C_{1:t}$ are the latest Temporal Statistics need to store for further use. Given the linear additivity of these statistics, we have:
\begin{align}
\small
\G_{\text{joint}}&=\!\!\!\!\!\!\sum_{(x,y)\in\train_{1:t}}\!\!\!\!\Phi(x)^\T\Phi(x)=\sum_{i=1}^{t}\!\sum_{(x,y)\in\train_{i}}\!\!\!\!\!\!\Phi(x)^\T\Phi(x)=\G_{1:t} \notag \\ \C_{\text{joint}}&=\!\!\!\!\sum_{(x,y)\in\train_{1:t}}\!\!\!\!\Phi(x)^\T\text{One-Hot}(y) \notag\\
&=\sum_{i=1}^{t}\!\sum_{(x,y)\in\train_{i}}\!\!\!\Phi(x)^\T\text{One-Hot}(y)=\C_{1:t}.
\label{joint}
\end{align}
This indicates that the statistics derived through Temporal Statistics Aggregation are consistently equivalent to those obtained in the joint setting (with access to all data from task $\mathcal{T}_1$ to task $\mathcal{T}_t$).

\noindent
\textbf{Classifier Update.} After the first stage adaptation, the backbone remains fixed, and only the classifier is updated. The objective of our \methodall{} is to build a strong classifier capable of accurately distinguishing all previously encountered classes. We formulate the solving procedure of the classifier as a regularized least squares form (ridge regression):
\begin{align}
\label{lp} 
\min_{\boldsymbol{W}} \|\boldsymbol{X}_\text{joint}\boldsymbol{W}-\boldsymbol{Y}_\text{joint}\|_\mathrm{F}^2+\gamma\|\boldsymbol{W}\|_\mathrm{F}^2.  
\end{align}
Here, $\X_\text{joint}$ represents the complete feature matrix, $\Y_\text{joint}$ represents the label matrix of all data samples from tasks $\mathcal{T}_1$ to $\mathcal{T}_t$, $\gamma$ controls the strength of the regularization. The close-formed solution of Eq. \ref{lp} is:
\begin{align}
\W^{\star}&=(\G_\text{joint}+\gamma\boldsymbol{I})^{-1}\C_\text{joint} 
\end{align}
Based on Eq. \ref{joint}, we have:
\begin{align}
\label{update_classifier}
\W_t&=(\G_{1:t}+\gamma\boldsymbol{I})^{-1}\C_{1:t} \notag\\
&=(\G_\text{joint}+\gamma\boldsymbol{I})^{-1}\C_\text{joint} =\W^{\star}.
\end{align}
Through our Spatial-Temporal Statistics Aggregation, we can obtain the feature statistics required for ridge regression on all data at any stage. Finally, the global model at the end of each stage $t$ is $\theta_t=\{F,R,\boldsymbol{W}_t\}$. 

\textbf{Discussion.} The rationale for STSA lies in the fact that the feature extractor is typically task-agnostic \cite{revisiting, cl_survey2, acil, ranpac, fecam, ease}. It learns to recognize "visual primitives", such as edges, textures. For example, an extractor that learns "wheels" and "metal frames" from the "car" class can combine these "primitives" to build a strong representation for a newly introduced "bicycle" class. Consequently, the incremental learning simplifies from re-optimizing the entire feature space to learning a new decision boundary within an already expressive feature space. Built on this, our \methodshort{} extends online least squares \cite{filter, robot, acil, ranpac, guanstatistics} to the FCIL scenario by providing a unified way to aggregate feature statistics across both temporal and spatial dimensions. In any scenario with restricted access to the complete dataset, the required statistics for ridge regression on the full dataset can be effectively obtained through our aggregation scheme.

\subsection{A communication-efficient variant (STSA-E)}
\label{STSA_E}
The communication overhead of uploading $\G_{t,k}$ ($M \times M$) increases significantly as $M$ grows. To address this, we propose a communication-efficient variant of \methodshort{}, named \methodshortvar{}. It enables the server to approximate global second-order feature statistics $\G_{t}$ using first-order statistics uploaded from clients.

\textbf{Proposition 3.1} \textit{For each task $\mathcal{T}_t$,
\begin{align}
\label{estimated_G}
\hat{\G_t} &= \sum_{i=1}^{c_t} \Bigg[ 
    \frac{n^{(i)}_t - 1}{K - 1} 
    \sum_{k=1}^K 
    \frac{\big( \C_{t,k}^{(i)}  \big) \big( \C_{t,k}^{(i)}  \big)^\top}{n_{t,k}^{(i)}} 
    \notag \\
    &- \frac{n^{(i)}_t - K}{n^{(i)}_t (K - 1)} 
    \bigg( \sum_{k=1}^K \C_{t,k}^{(i)}  \bigg) 
    \bigg( \sum_{k=1}^K \C_{t,k}^{(i)}  \bigg)^\top 
\Bigg]
\end{align}
is an unbiased plug-in estimator of the global gram matrix $\G_t$ when the global feature set \text{~$\Omega=\{(\Phi(x),y)\mid(x,y)\in \cup_{k=1}^K \train_{t,k}\}$} is i.i.d. Here, $\C_{t,k}^{(i)}$ represents the $i$-th column of $\C_{t,k}$, $n_t^{(i)}$ is the total number of samples from class $i$ at stage $t$, and $n_{t,k}^{(i)}$ is the number of samples from class $i$ for client $k$ at stage $t$.
\label{G}
}

\textbf{Remark 1.} The i.i.d. assumption of the global feature set $\Omega$ in Proposition 3.1 is valid within the non-i.i.d. setting (data heterogeneity) in FCIL. We explain it in \textbf{Appendix \ref{proof_iid}}.

\vspace{-3pt}
Proposition 3.1 indicates that each client $k$ only needs to transmit the correlation matrix $\C_{t,k}$ and the label frequency vector $\nn_{t,k}$ ($\nn_{t,k} \in \R^{c_t}$) for task $t$, significantly reducing communication overhead. The server leverages these uploaded statistics to compute the estimated gram matrix $\hat{\G_t}$ and performs temporal statistics aggregation (Eq. \ref{temporal_aggregation}) to derive the accumulated global gram matrix $\hat{\G}_{1:t}$. Then $\G_{1:t}$ is used in Eq. \ref{update_classifier} to update the classifier:
\begin{align}
\label{update_classifier2}
    \boldsymbol{W}_t=(\hat{\G}_{1:t}+\gamma\boldsymbol{I})^{-1}\C_{1:t}.
\end{align}
Then we provide two key theoretical results to analyze the estimation error of \methodshortvar{} in comparison to \methodshort{}. The full descriptions and proofs can be found in \textbf{Appendix \ref{proof_stsa_e}}.

We consider a noisy linear regression model, where $n=\sum_{i=1}^t n_i$ and $n>M$. Specifically, we assume the existence of a ground-truth weight matrix $\boldsymbol{W}_t^0 \in \mathbb{R}^{M\times c_{\text{~joint}}}$ and noise $\mathcal{E}_{joint} \in \mathbb{R}^{n\times c_{\text{~joint}}}$ such that $\boldsymbol{Y}_{\text{~joint}}= \boldsymbol{X}_{\text{~joint}}\boldsymbol{W}_t^0+\mathcal{E}_{\text{~joint}}$, where $c_{\text{~joint}}=\sum_i^t c_i$. 

For each task $\mathcal{T}_t$, and for every $i \in\left\{1, \ldots, c_t\right\}$, we make the following assumptions:

\textbf{Assumption 1 (Data Structure).} $\|\boldsymbol{X}_{t}\|_F\leq \sqrt{n_t} C_{\boldsymbol{X}}$, $\operatorname{tr}\left(\Sigma_t^{(i)}\right) \leq C_{\Sigma}$ and $\|\mu_t^{(i)}\|_2^2 \leq C_{\mu}$ for some constants $C_{\boldsymbol{X}},C_{\Sigma}, C_{\mu}>0$, where $\mu_t^{(i)}$ and $\Sigma_t^{(i)}$ are the class-wise mean and covariance of the global feature set $\Omega$ at stage $t$.

\textbf{Assumption 2 (Class Balance).} There exists a constant $0<C\ll c_t$ such that $c_t\sum_{i=1}^{c_t} n_t^{(i)2}\leq Cn_t^2$.

\textbf{Assumption 3 (Stability).} $\|\boldsymbol{W}_t^{0}\|_F \leq \sqrt{M}r$ for some constant $r>0$.

\textbf{Assumption 4 (Noise).} The $i$-th row of noise satisfies the following conditions: $\mathbb{E}(\mathcal{E}_{\text{~joint},i}^\top)=\mathbf{0}$, $\mathbb{E}\left(\mathcal{E}_{\text{~joint},i}^{\top} \mathcal{E}_{\text{~joint},i}\right)=\sigma^2 \boldsymbol{I}_{M}$ and $\mathcal{E}_{\text{~joint}}$ is independent of $\boldsymbol{X}_{\text{~joint}}$.

These four assumptions are mild in statistical learning theory and aim to reflect practical conditions without overidealization. We provide a detailed explanation of their rationality in \textbf{Appendix \ref{proof_stsa_e}}.

\textbf{Theorem 3.3 (\methodshortvar{} plug-in error).}
\textit{Under Assumptions 1-2, with high probability, we have
\begin{align}
    \frac{1}{n}\mathbb{E}\left\|\boldsymbol{X}_{\text{~joint}}\boldsymbol{W}_t-\boldsymbol{X}_{\text{~joint}}\boldsymbol{W}^{\star}\right\|_F\leq O\left(\frac{\frac{K+1}{K-1}t}{n(\frac{\gamma^2}{n^2}+1)}\right).
\end{align}
}

\noindent
\textbf{Remark 2.} Theorem 3.3 shows that the plug-in error decreases as the average per-stage sample size $\frac{n}{t}$ increases.

\textbf{Corollary 3.4 (Estimation error).}
\textit{Under Assumptions 1-4, with high probability, we have
\begin{align}
    \frac{1}{n}\mathbb{E}&\left\|\boldsymbol{X}_{\text{~joint}}\boldsymbol{W}_t-\boldsymbol{Y}_{\text{~joint}}\right\|_F \leq O\left(\frac{\frac{K+1}{K-1}t}{n(\frac{\gamma^2}{n^2}+1)}\right)+O\left(\frac{1}{\sqrt{n}}\right).
\end{align}
}

\noindent
\textbf{Remark 3.} Corollary 3.4 reveals that our \methodshortvar{} achieves $O\left(\mathrm{tn}^{-1}\right)$ estimation efficiency, contrasting with the oracle estimator's $O\left(n^{-1 / 2}\right)$ rate (Eq. \ref{update_classifier}) \citep{hsu2012random}. Thus, \methodshortvar{} maintains comparable statistical performance to the \methodshort{} with significantly reduced communication overhead, achieving a great performance-resource trade-off.
\vspace{-5pt}

\noindent
\textbf{Dummy Clients.} Since the number of clients, $K$, affects the quality of the estimated $\hat{\G_t}$, we introduce dummy clients to improve estimation when $K$ is small (cross-silo setting). Specifically, we divide each client's local dataset into multiple subsets, treating each subset as a separate dummy client. For instance, client $k$ in task $\mathcal{T}_t$ splits its dataset $\train_{t,k}$ into $K_D$ subsets. Each dummy client $k_j$ ($j = 1, 2, \ldots, K_D$) holds a subset $\train_{t,k_j} \subset \train_{t,k}$ and uploads the corresponding spatial statistics $\C_{t,k_j}$ and $\nn_{t,k_j}$ to the server. This effectively increases the number of clients from $K$ to $K \times K_D$.
\textbf{Discussion.} Introducing dummy clients brings extra communication overhead for uploading first-order statistics (from $\C_{t,k}$ to $\{\C_{t,k_j}, \nn_{t,k_j}\}_{j=1}^{K_D}$). However, this overhead is typically much smaller than the cost of uploading second-order statistics ($\G_{t,k}$) in STSA (see Section \ref{sec:further}). It is worth noting that dummy clients are used to adapt \methodshortvar{} to the cross-silo setting, where the number of clients $K$ is small. In the cross-device setting ($K$ is large), they are not needed (see Section \ref{sec:ablation}).

\vspace{-5pt}
Figure \ref{fig:framework} shows an overview of \methodall{} and Algorithm \ref{alg:USTA} provides its entire procedure with pseudocode.

\vspace{-10pt}
\section{Experiments}

\begin{table*}[t]
\vspace{-8pt}
\caption{Experimental results for training from scratch (ResNet18) with varying levels of data heterogeneity. The best and second-best results are shown \textbf{in bold} and \underline{underlined}, respectively.} 
\vspace{3pt}
\renewcommand{\arraystretch}{1.1} 
\begin{adjustbox}{width=1.0\linewidth}{
\begin{tabular}{c|lccccccccc}
\hline
\multirow{2}{*}{Dataset} & Data Partition & \multicolumn{3}{c}{$\beta=1$} & \multicolumn{3}{c}{$\beta=0.5$} & \multicolumn{3}{c}{$\beta=0.1$} \\ \cline{3-11} 
                          & Metrics        & $A_{\text{avg}}$ (↑) & $A_T$ (↑) & $F_T$ (↓) & $A_{\text{avg}}$ (↑) & $A_T$ (↑) & $F_T$ (↓) & $A_{\text{avg}}$ (↑) & $A_T$ (↑) & $F_T$ (↓) \\ \hline
\multirow{6}{*}{CIFAR100} 
& Joint (Upper-bound)  & - & 76.74 & - & - & 76.74 & - & - & 76.74 & - \\ \cline{2-11} 
& Fed-EWC \cite{forget2} & 29.23 & 11.76 & 73.52 & 27.16 & 11.56 & 70.53 & 21.21 & 9.13 & 60.79 \\
                          & Fed-LWF \cite{lwf}     & 44.89 & 27.04 & 44.60 & 40.25 & 24.23 & 42.86 & 32.28 & 21.89 & 38.29 \\
                          & MFCL \cite{mfcl}       & 43.48 & 26.46 & 31.28 & 44.00 & 27.17 & 27.27 & 34.48 & 21.96 & 22.77 \\
                          & LANDER \cite{lander}   & \textbf{55.01} & \textbf{37.53} & 28.31 & 45.68 & 33.09 & 35.38 & 42.41 & 30.50 & 26.42 \\ \cline{2-11} 
                        & \cellcolor{Thistle!20}\methodshortvar{} (Ours) & \cellcolor{Thistle!20}50.29 & \cellcolor{Thistle!20}34.85 & \cellcolor{Thistle!20}\textbf{9.69} & \cellcolor{Thistle!20}\underline{49.16} & \cellcolor{Thistle!20}\underline{34.57} & \cellcolor{Thistle!20}\textbf{9.26} & \cellcolor{Thistle!20}\underline{44.86} & \cellcolor{Thistle!20}\underline{30.74} & \cellcolor{Thistle!20}\underline{9.92} \\
                        & \cellcolor{Thistle!20}\methodshort{} (Ours)    & \cellcolor{Thistle!20}\underline{50.71} & \cellcolor{Thistle!20}\underline{35.14} & \cellcolor{Thistle!20}\underline{9.81} & \cellcolor{Thistle!20}\textbf{50.10} & \cellcolor{Thistle!20}\textbf{34.97} & \cellcolor{Thistle!20}\underline{9.57} & \cellcolor{Thistle!20}\textbf{46.19} & \cellcolor{Thistle!20}\textbf{31.02} & \cellcolor{Thistle!20}\textbf{9.73} \\ \hline
\multirow{6}{*}{Tiny-ImageNet} 
& Joint (Upper-bound)  & - & 49.44 & - & - & 49.44 & - & - & 49.44 & - \\ \cline{2-11} 
& Fed-EWC \cite{forget2} & 19.49 & 7.68 & 52.51 & 19.07 & 7.48 & 52.04 & 17.55 & 6.08 & 49.82 \\
                               & Fed-LWF \cite{lwf}     & 29.11 & 17.02 & 27.64 & 26.91 & 15.47 & 28.06 & 26.54 & 15.16 & 25.22 \\
                               & MFCL \cite{mfcl}       & \underline{32.58} & 18.12 & 24.53 & 30.35 & 16.43 & 28.93 & 28.21 & 16.24 & 36.33 \\
                               & LANDER \cite{lander}   & \textbf{34.53} & 18.21 & 38.04 & 30.75 & 16.59 & 26.77 & 28.49 & 16.29 & 33.76 \\ \cline{2-11} 
                            & \cellcolor{Thistle!20}\methodshortvar{} (Ours) & \cellcolor{Thistle!20}30.86 & \cellcolor{Thistle!20}\underline{19.94} & \cellcolor{Thistle!20}\underline{8.94} & \cellcolor{Thistle!20}\textbf{30.91} & \cellcolor{Thistle!20}\textbf{20.02} & \cellcolor{Thistle!20}\underline{7.66} & \cellcolor{Thistle!20}\underline{29.07} & \cellcolor{Thistle!20}\underline{18.11} & \cellcolor{Thistle!20}\underline{8.36} \\
                            & \cellcolor{Thistle!20}\methodshort{} (Ours)    & \cellcolor{Thistle!20}31.14 & \cellcolor{Thistle!20}\textbf{20.03} & \cellcolor{Thistle!20}\textbf{8.26} & \cellcolor{Thistle!20}\underline{30.77} & \cellcolor{Thistle!20}\underline{19.99} & \cellcolor{Thistle!20}\textbf{6.56} & \cellcolor{Thistle!20}\textbf{29.10} & \cellcolor{Thistle!20}\textbf{18.18} & \cellcolor{Thistle!20}\textbf{7.50} \\ \hline
\end{tabular}
}\end{adjustbox} 
\label{scratch}
\vspace{-15pt}
\end{table*}


\begin{table*}[t]
\caption{Experimental results for training from pre-trained model (ViT-B/16) with varying levels of data heterogeneity. The best and second-best results are shown \textbf{in bold} and \underline{underlined}, respectively.}
\vspace{3pt}
\renewcommand{\arraystretch}{1.1} 
\begin{adjustbox}{width=1.0\linewidth}{
\begin{tabular}{c|lccccccccc}
\hline
\multirow{2}{*}{Dataset} & Data Partition & \multicolumn{3}{c}{$\beta=1$} & \multicolumn{3}{c}{$\beta=0.5$} & \multicolumn{3}{c}{$\beta=0.1$} \\ \cline{3-11} 
                          & Metrics        & $A_{\text{avg}}$ (↑) & $A_T$ (↑) & $F_T$ (↓) & $A_{\text{avg}}$ (↑) & $A_T$ (↑) & $F_T$ (↓) & $A_{\text{avg}}$ (↑) & $A_T$ (↑) & $F_T$ (↓) \\ \hline
\multirow{6}{*}{CIFAR100} 
& Joint (Upper-bound) & - & 92.45 & - & - & 92.45 & - & - & 92.45 & - \\ \cline{2-11} 
& Fed-DualP \cite{dualprompt} & 83.02 & 75.07 & 4.78 & 78.12 & 65.77 & \textbf{3.97} & 69.11 & 56.31 & \textbf{3.31} \\
                          & Fed-CODAP \cite{coda}       & 82.20 & 74.12 & 6.34 & 74.80 & 64.02 & 10.79 & 67.75 & 55.54 & 14.04 \\
                          & Fed-CPrompt \cite{cprompt}  & 81.43 & 73.86 & 8.29 & 75.07 & 65.87 & 9.28 & 69.67 & 61.68 & 10.03 \\
                          & PILoRA \cite{pilora}        & 83.59 & 77.17 & 8.59 & 83.46 & 76.81 & 6.41 & 82.85 & 74.09 & 7.67 \\ \cline{2-11} 
                        & \cellcolor{Thistle!20}\methodshortvar{} (Ours) & \cellcolor{Thistle!20}\underline{93.71} & \cellcolor{Thistle!20}\underline{89.42} & \cellcolor{Thistle!20}\textbf{3.93} & \cellcolor{Thistle!20}\underline{93.35} & \cellcolor{Thistle!20}\underline{89.37} & \cellcolor{Thistle!20}\underline{4.05} & \cellcolor{Thistle!20}\underline{93.14} & \cellcolor{Thistle!20}\underline{88.65} & \cellcolor{Thistle!20}\underline{4.06} \\
                        & \cellcolor{Thistle!20}\methodshort{} (Ours)    & \cellcolor{Thistle!20}\textbf{94.05} & \cellcolor{Thistle!20}\textbf{89.85} & \cellcolor{Thistle!20}\underline{4.12} & \cellcolor{Thistle!20}\textbf{93.96} & \cellcolor{Thistle!20}\textbf{89.72} & \cellcolor{Thistle!20}4.09 & \cellcolor{Thistle!20}\textbf{93.63} & \cellcolor{Thistle!20}\textbf{89.09} & \cellcolor{Thistle!20}4.35 \\ \hline
\multirow{6}{*}{ImageNet-R} 
& Joint (Upper-bound)  & - & 80.27 & - & - & 80.27 & - & - & 80.27 & - \\ \cline{2-11} 
& Fed-DualP \cite{dualprompt} & 56.61 & 48.58 & 7.13 & 53.37 & 45.25 & 9.80 & 35.58 & 27.92 & 10.52 \\

                          & Fed-CODAP \cite{coda}       & 55.10 & 51.13 & \textbf{6.42} & 48.49 & 43.00 & \textbf{6.06} & 23.22 & 15.15 & 8.29 \\
                          & Fed-CPrompt \cite{cprompt}  & 51.09 & 47.28 & 9.49 & 45.42 & 39.12 & 9.27 & 23.67 & 16.87 & 9.52 \\
                          & PILoRA \cite{pilora}        & 55.47 & 51.92 & 6.69 & 53.43 & 50.18 & 7.09 & 51.85 & 47.85 & 7.19 \\ \cline{2-11} 
                        & \cellcolor{Thistle!20}\methodshortvar{} (Ours) & \cellcolor{Thistle!20}\underline{72.89} & \cellcolor{Thistle!20}\underline{66.68} & \cellcolor{Thistle!20}\underline{6.45} & \cellcolor{Thistle!20}\underline{71.77} & \cellcolor{Thistle!20}\underline{66.08} & \cellcolor{Thistle!20}\underline{6.60} & \cellcolor{Thistle!20}\textbf{71.67} & \cellcolor{Thistle!20}\underline{65.46} & \cellcolor{Thistle!20}\textbf{6.37} \\
                        & \cellcolor{Thistle!20}\methodshort{} (Ours)    & \cellcolor{Thistle!20}\textbf{74.02} & \cellcolor{Thistle!20}\textbf{67.25} & \cellcolor{Thistle!20}6.84 & \cellcolor{Thistle!20}\textbf{73.59} & \cellcolor{Thistle!20}\textbf{66.67} & \cellcolor{Thistle!20}7.14 & \cellcolor{Thistle!20}\textbf{73.65} & \cellcolor{Thistle!20}\textbf{66.78} & \cellcolor{Thistle!20}\underline{7.18} \\ \hline
\end{tabular}
}\end{adjustbox} 
\label{pre-trained}
\vspace{-10pt}
\end{table*}

\subsection{Experimental Setting}
\label{exp:setting}
\noindent
\textbf{Datasets and Models.} 
We conduct experiments under two settings: training from scratch and training with a pre-trained model. For the scratch setting, we use ResNet18 \cite{resnet} as the backbone on CIFAR100 \cite{c10} and Tiny-ImageNet \cite{tin}. For the pre-trained setting, we use ViT-B/16 \cite{vit} (pre-trained on ImageNet-21K) as the backbone on CIFAR100 \cite{c10} and ImageNet-R \cite{imagenetr}. By default, we divide dataset classes into 10 non-overlapping tasks ($T\!\!=\!\!10$) and distribute them among 5 clients ($K\!\!=\!\!5$) using a Dirichlet distribution with varying $\beta$.

\vspace{-3pt}
\noindent
\textbf{Comparison Baselines.} 
Following the benchmark in \cite{target, lander}, we compare our \methodshort{} and \methodshortvar{} with Fed-EWC \cite{forget2}, Fed-LWF \cite{lwf}, MFCL \cite{mfcl}, and LANDER \cite{lander} under the training from scratch setting. We also evaluate our method against PEFT-based FCIL methods, including Fed-DualP \cite{dualprompt}, Fed-CODAP \cite{coda}, Fed-CPrompt \cite{cprompt}, and PILoRA \cite{pilora}, using a pre-trained ViT-B/16. For our method, we integrate an adapter \cite{revisiting} with ViT to ensure performance after the first task is comparable to other PEFT-based methods.

\vspace{-3pt}
\noindent
\textbf{Evaluation Metrics.} The evaluation follows traditional CIL metrics \cite{coda, dualprompt, revisiting}, including: (1) average incremental accuracy $A_{\text{avg}}$, (2) final average accuracy $A_T$ and (3) average forgetting $F_T$. Refer to \textbf{Appendix \ref{appendix:metrics}} for further details on these metrics.

\vspace{-3pt}
\noindent
\textbf{Experimental Details.} For ResNet18, we set the batch size to 128, local training epoch to 2, and communication rounds to 100. For pre-trained ViT-B/16, we use 2 local epochs with a batch size of 128, conducting 10 communication rounds on CIFAR100 and 20 communication rounds on ImageNet-R. The projection dimension of the adapter is set to 64. The dimension of the random feature $M$ is set to 5000 for ResNet18 and 1250 for ViT-B/16. The number of dummy clients $K_D$ is set to 50 for ResNet18, 10 for ViT-B/16. The ridge coefficient is set to 1e4 for ResNet18 and 1e6 for ViT-B/16. More details on configurations can be found in \textbf{Appendix \ref{appendix:client_training}}. 

\vspace{-10pt}
\subsection{Main Results}
\begin{figure*}[h]
\begin{center}

\includegraphics[width=1.0\linewidth]{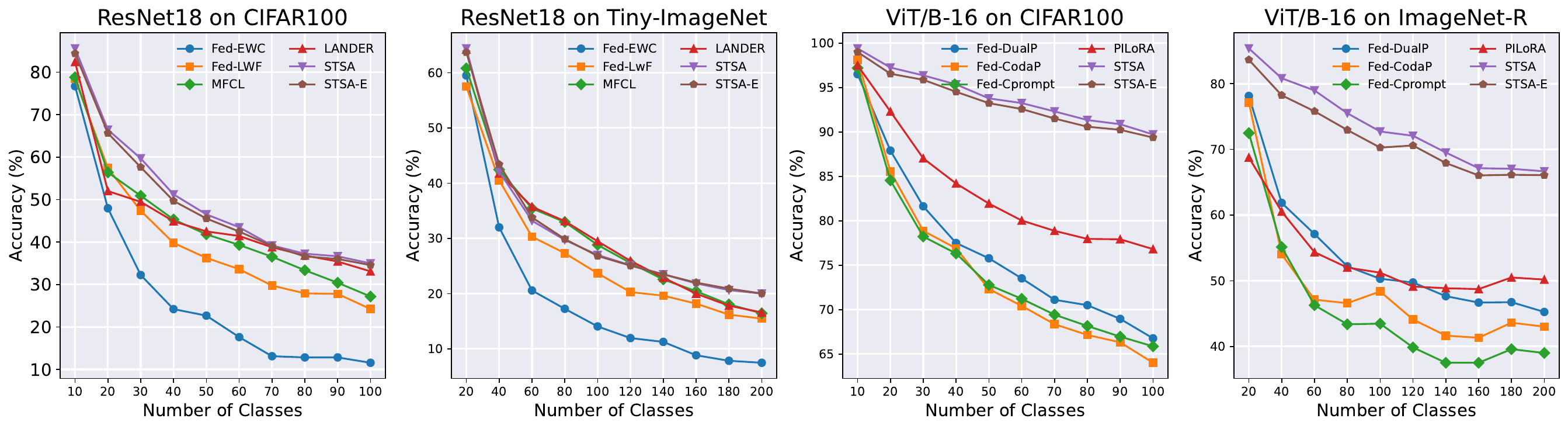}
\end{center}
\vspace{-15pt}
\caption{Performance curve of different datasets ($\beta=0.5$) with different backbone.}
\label{fig:pc_c100}
\vspace{-15pt}
\end{figure*}

\noindent
\textbf{Experiments for training from scratch.} We present the results of experiments using ResNet18 in Table \ref{scratch} and Figure \ref{fig:pc_c100}. When data heterogeneity is low ($\beta=1$), the DFKD method (LANDER, MFCL) performs well, and our method is slightly weaker than the best baseline, LANDER. However, as data heterogeneity increases, our method demonstrates greater robustness. On CIFAR100, \methodall{} shows significant performance improvements, particularly in average incremental accuracy. At $\beta=0.5$, \methodshortvar{} outperforms LANDER by 3.48\%, and \methodshort{} surpasses LANDER by 4.42\%. Similarly, on Tiny-ImageNet, \methodall{} shows substantial gains in final incremental accuracy. When $\beta=0.5$, \methodshortvar{} outperforms LANDER by 3.43\%, and \methodshort{} surpasses LANDER by 3.4\%. Under extreme heterogeneity ($\beta=0.1$), our method, like others, shows some performance drop due to the less effective feature extractor trained in the first stage. This drop can be mitigated by using better federated learning algorithms or a pre-trained model in the first stage.

\vspace{-5pt}
\noindent
\textbf{Experiments for training from pre-trained model.} The experimental results of training from pre-trained ViT-B/16 are shown in Table \ref{pre-trained}. We observe that \methodall{} and \methodshortvar{} significantly outperforms the best FCIL baseline, achieving over \textbf{10\%} improvements in both average incremental accuracy and final average accuracy across all cases. For CIFAR100, our method is just 3\% lower than the upper bound. Furthermore, \methodall{} and \methodshortvar{} exhibit strong robustness to data heterogeneity, maintaining nearly consistent performance regardless of the degree of heterogeneity. Moreover, we show the performance curve of different methods in Figure \ref{fig:pc_c100}, which highlights the outstanding performance of our method throughout the entire incremental learning process. Compared to training from scratch, the pre-trained model's generalizability enables it to learn a strong feature extractor in the first phase, which gives our method a significant performance advantage.

\noindent
\textbf{Comparison between STSA and STSA-E.}
STSA typically slightly outperforms STSA-E, though the reverse can occur, such as on Tiny-ImageNet ($\beta=0.5$). This is due to estimation errors, which can be seen as random noise that sometimes enhances performance. As noted in Appendix \ref{privacy_tech}, adding random noise to feature statistics can sometimes improve performance. Similar findings have been reported in several FL studies \cite{dbe, zhang2024enhancing}. In summary, STSA-E delivers comparable performance to STSA while significantly reducing communication overhead (see Section \ref{sec:further}). This makes STSA-E particularly well-suited for communication-constrained environments, such as mobile and IoT applications.

\vspace{-5pt}
\subsection{Ablation Study}
\label{sec:ablation}
\vspace{-5pt}
We conduct the ablation study on the CIFAR100 with $\beta = 0.5$. Results for additional settings can be found in \textbf{Appendix \ref{full_ablation}}.

\begin{figure}[h]
 \centering
\includegraphics[width=1.0\linewidth]{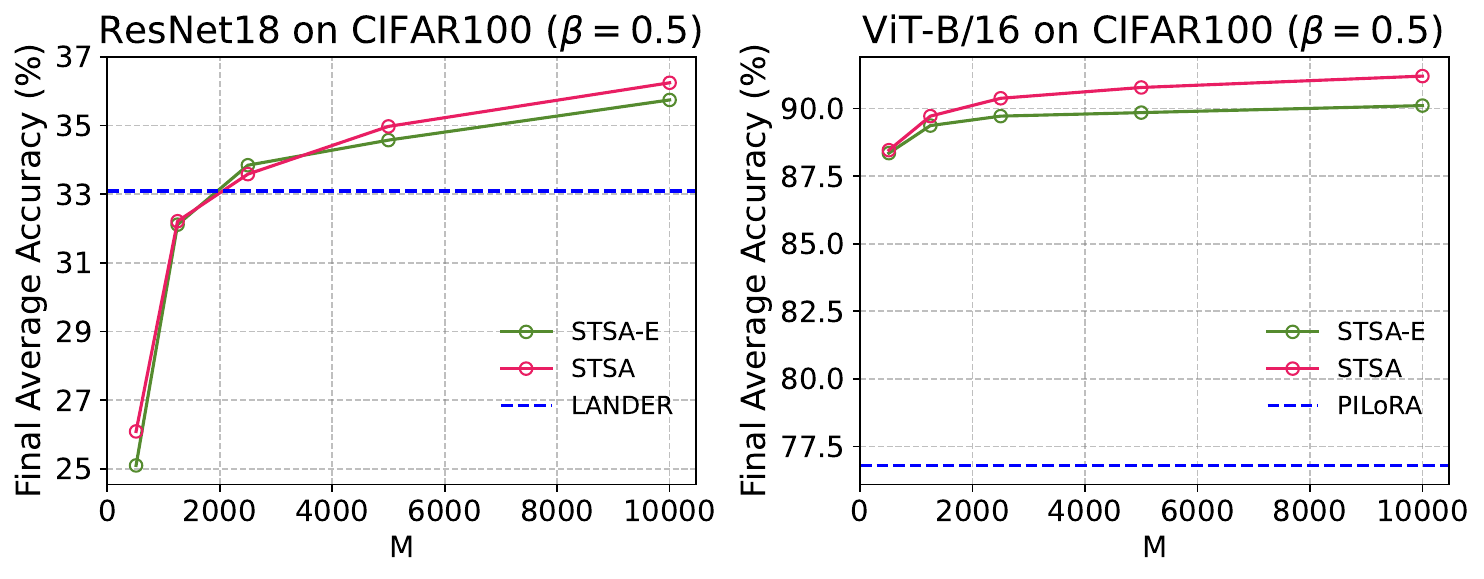}
\vspace{-20pt}
\caption{Final average accuracy $A_{T}$ with different $M$. The top-performing baselines (LANDER and PILoRA) are also marked in the figure.}
\vspace{-10pt}
\label{fig:M}
\end{figure}

\noindent
\textbf{Ablation on Dimension of Random Feature.} Figure \ref{fig:M} shows the results with different values of $M$. Here, 512 is the dimension of raw feature in ResNet18, and 768 is the dimension of raw feature in ViT-B/16. We find that random mapping significantly improves the performance of \methodall{}, confirming that random mapping enhances the linear separability of features. Additionally, setting $M$ above 1250 is sufficient to achieve competitive results compared to other state-of-the-art baselines.
\vspace{-5pt}

\begin{figure}[h]
 \centering
\includegraphics[width=1.0\linewidth]{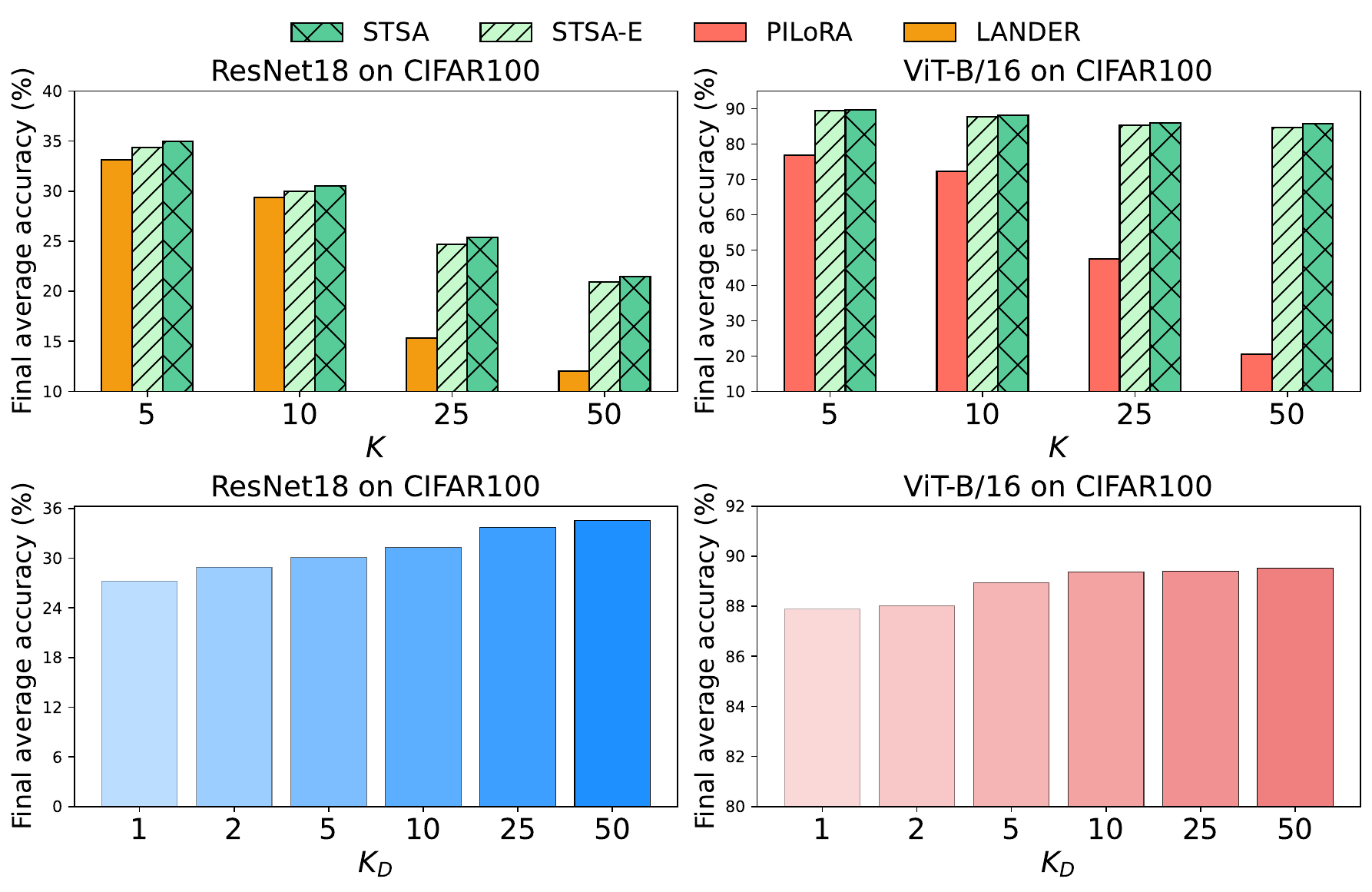}
\vspace{-20pt}
\caption{\textbf{(Top)} Final average accuracy with varying client number $K$. The top-
performing baselines (LANDER and PILoRA) are also marked in the figure. \textbf{(Bottom)} Final average accuracy with varying dummy client number $K_D$.}
\label{fig:client_num}
\vspace{-10pt}
\end{figure}

\noindent
\textbf{Ablation on Number of Clients.} 
The number of clients varied as $K = \{5, 10, 25, 50\}$. For \methodshortvar{}, we set $K_D$ to $\{50, 25, 10, 5\}$ for ResNet18, totaling 250 clients ($K \times K_D$), and $\{10, 5, 2, 1\}$ for ViT-B/16, totaling 50 clients. Figure \ref{fig:client_num} (Top) demonstrates that our method outperforms other top-performing baselines as the number of clients increases. Notably, performance advantage becomes more evident with more clients, as only the first-stage adaptation is influenced by the number of clients, while the subsequent spatial-temporal statistics aggregation remains unaffected. Meanwhile, when the number of clients is large, there is no need to set dummy clients (ViT with $K=50,K_D=1$). Therefore, \methodshortvar{} is particularly well-suited for cross-device settings with many clients. We also assess the impact of varying the number of dummy clients $K_D$ on the performance of \methodshortvar{}. As shown in Figure \ref{fig:client_num} (Bottom), the performance of \methodshortvar{} improves with more dummy clients and gradually approaches that of \methodshort{}.

\noindent
\textbf{Ablation on First Stage adaptation Strategies.} 
The subsequent statistics aggregation is independent of the first stage adaptation, so our \methodshort{} and  \methodshortvar{} can combine with any PEFT methods for the first stage. Here, we combine two PEFT-based FCIL methods, PILoRA (LoRA) and Fed-CPrompt (prompt), with ours. Table \ref{table:first_stage} shows the results. We find that our \methodall{} consistently achieves better performance against other baselines in all metrics. This demonstrates the high flexibility of our method.

\vspace{-10pt}
\begin{table}[h]
\centering
\begin{adjustbox}{width=1.0\linewidth}
\begin{tabular}{l|lccc}
\toprule
PEFT & Metrics & $\small A_{\text{avg}}$(↑) & $A_T$(↑) & $F_T$(↓)  \\
\midrule
\multirow{3}{*}{LoRA} & PILoRA \cite{pilora} & 83.46 & 76.81 & 6.41 \\
& \cellcolor{Thistle!20}\methodshortvar{}+PILoRA & \cellcolor{Thistle!20}\underline{89.45} & \cellcolor{Thistle!20}\underline{84.72} & \cellcolor{Thistle!20}\textbf{5.54} \\
& \cellcolor{Thistle!20}\methodshort{}+PILoRA      & \cellcolor{Thistle!20}\textbf{90.51} & \cellcolor{Thistle!20}\textbf{85.64} & \cellcolor{Thistle!20}\underline{5.82} \\
\midrule
\multirow{3}{*}{Prompt} &
Fed-CPrompt \cite{cprompt}  & 75.07 & 65.87 & 9.28 \\
& \cellcolor{Thistle!20}\methodshortvar{}+Fed-CPrompt & \cellcolor{Thistle!20}\textbf{90.07} & \cellcolor{Thistle!20}\textbf{85.67} & \cellcolor{Thistle!20}\textbf{5.26} \\
& \cellcolor{Thistle!20}\methodshort{}+Fed-CPrompt  & \cellcolor{Thistle!20}89.86 & \cellcolor{Thistle!20}\underline{85.60} & \cellcolor{Thistle!20}\underline{6.32} \\
\midrule
\multirow{2}{*}{Adapter} &
\cellcolor{Thistle!20}\methodshortvar{}    & \cellcolor{Thistle!20}\underline{93.35} & \cellcolor{Thistle!20}\underline{89.37} & \cellcolor{Thistle!20}\textbf{4.05} \\
& \cellcolor{Thistle!20}\methodshort{}     & \cellcolor{Thistle!20}\textbf{93.96} & \cellcolor{Thistle!20}\textbf{89.72} & \cellcolor{Thistle!20}\underline{4.09} \\ 
\bottomrule
\end{tabular}
\end{adjustbox}
\caption{Results of different first stage adaptation strategies. The dataset is CIFAR100 ($\beta=0.5$).}
\label{table:first_stage}
\vspace{-15pt}
\end{table}

\subsection{Further Analysis}
\label{sec:further}
\noindent
\textbf{Large First Task Setting.} The performance of our method relies on the feature extractor trained during the first incremental stage. A larger initial dataset improves the extractor's quality. To examine this, we conduct experiments where the first task includes 50\% of the total classes. Specifically, we partition the CIFAR100 dataset into 11 tasks: the base task (denoted as \( t = 0 \)) contains 50 classes, while each subsequent task (for \( t > 0 \)) consists of 5 classes. Table \ref{tab:big_task} presents the results. Our method consistently shows strong performance, outperforming top baselines by nearly 10\% in all cases. This indicates that our method has the potential to achieve much better performance if the backbone trained in the first stage is strong enough.

\vspace{-7pt}
\begin{table}[H]
\centering
\begin{adjustbox}{width=1.0\linewidth}
\begin{tabular}{c|lccc}
\toprule
Backbone & Metrics & $\small A_{\text{avg}}$(↑) & $A_T$(↑) & $F_T$(↓)  \\
\midrule
\multirow{3}{*}{ResNet18} & LANDER \cite{lander}& 59.69  & 46.19 & 13.61 \\
                        & \cellcolor{Thistle!20}\methodshortvar{} (Ours)& \cellcolor{Thistle!20}\underline{66.78}  & \cellcolor{Thistle!20}\underline{59.36} & \cellcolor{Thistle!20}\underline{3.23}  \\
                        & \cellcolor{Thistle!20}\methodshort{} (Ours)& \cellcolor{Thistle!20}\textbf{67.08}  & \cellcolor{Thistle!20}\textbf{59.44} & \cellcolor{Thistle!20}\textbf{2.69}  \\
                        \bottomrule  
\multirow{3}{*}{ViT-B/16}    & PILoRA \cite{pilora}& 86.24  & 81.44 & 4.21  \\
                        & \cellcolor{Thistle!20}\methodshortvar{} (Ours) & \cellcolor{Thistle!20}\underline{93.04}  & \cellcolor{Thistle!20}\underline{91.04} & \cellcolor{Thistle!20}\textbf{2.14}  \\
                        & \cellcolor{Thistle!20}\methodshort{} (Ours) & \cellcolor{Thistle!20}\textbf{93.70}  & \cellcolor{Thistle!20}\textbf{91.57} & \cellcolor{Thistle!20}\underline{2.19} \\
                        \bottomrule
\end{tabular}
\end{adjustbox}
\vspace{-10pt}
\caption{Comparison with top-performing baselines in Large First Task Setting.}
\vspace{-10pt}
\label{tab:big_task}
\end{table}

\begin{figure}[t]
 \centering
 \vspace{-2pt}
\includegraphics[width=1.0\linewidth]{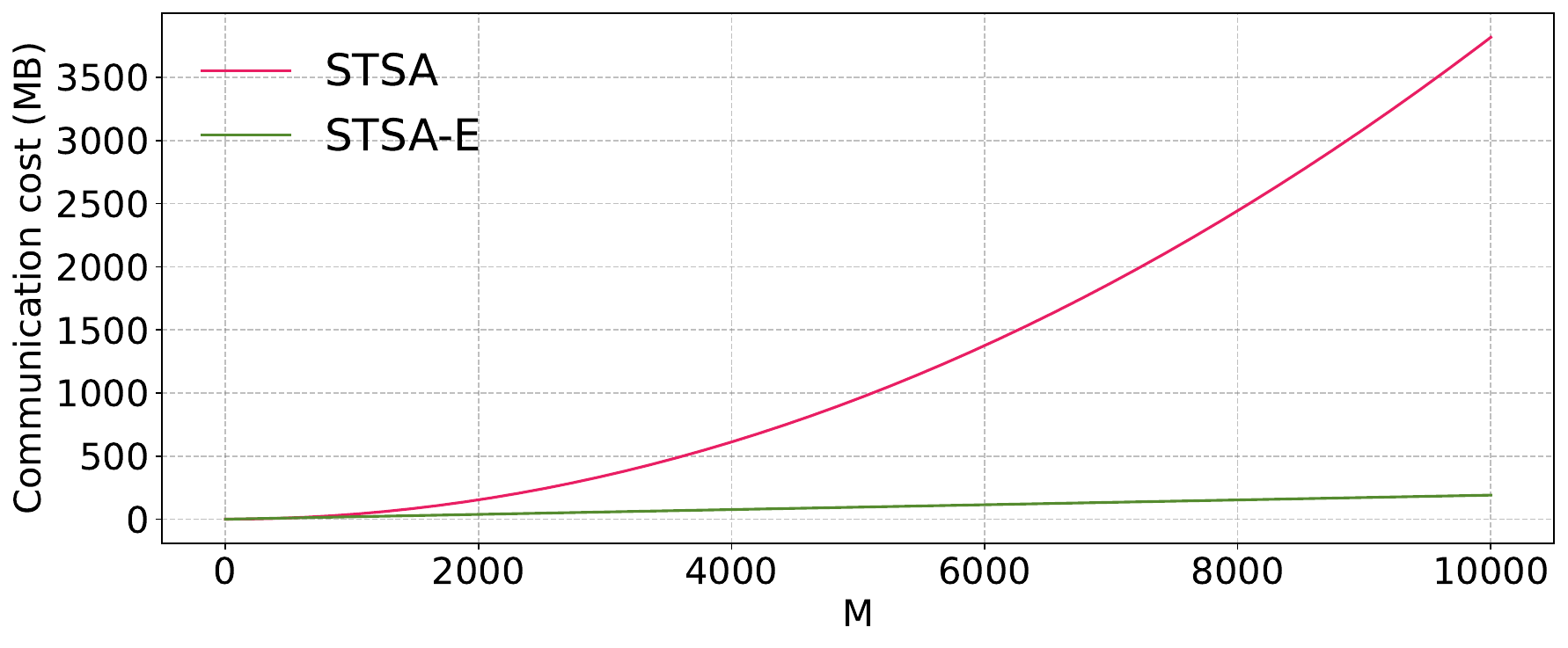}
\vspace{-25pt}
\caption{Communication overhead per client after the first stage of training, evaluated with varying $M$ on CIFAR100 ($K_D=50$).}

\label{fig:cost_curve}
\vspace{-10pt}
\end{figure}

\vspace{-8pt}
\noindent
\textbf{Communication Overhead.} After first stage adaptation, each client uploads only the Spatial Statistics, represented as $\{\G_{t,k}, \C_{t,k}\}$ with a size of $(M\!\!+\!c_t)\!\times\!M$ for \methodshort{}, or $\{\C_{t,k_j}, \nn_{t,k_j}\}_{j=1}^{K_D}$ with a size of $(M\!+\!1)\!\times\!c_t\!\times\!K_D$ for \methodshortvar{} at stage $t$. It is especially commendable that only one round of communication happens between the clients and the central server in the subsequent aggregation procedure, which further reduces the potential risk for man-in-the-middle attacks \cite{guan2025capture, coboosting}. The left column of Table \ref{tab:cost_combined} presents the communication overhead of the entire learning process for each client. For fair comparison, we only evaluate communication overhead after the training in the first stage. Our method significantly reduces communication overhead compared to DFKD methods and achieves results comparable to PEFT methods. Figure \ref{fig:cost_curve} shows that our \methodshortvar{} significantly reduces communication overhead compared to \methodshort{}.

\vspace{-3pt}
\noindent
\textbf{Computation Overhead.} Our \methodall{} does not involve the time-consuming backpropagation after the first stage adaptation, which provides a significant advantage in computational efficiency. We measure the total experimental time (including training and testing) after first stage adaptation using a single NVIDIA 3090 RTX GPU. As shown in the right columns in Table \ref{tab:cost_combined}, our \methodall{} is over 80 times faster than LANDER \cite{lander} and 120 times faster than PILoRA \cite{pilora}.

\vspace{-3pt}
\noindent
\textbf{Privacy of \methodall{}.} As same as PILoRA \cite{pilora}, our method involves uploading feature statistics, which may raise privacy concerns. Specifically, \methodshort{} uploads $\G_{t,k}$ and $\C_{t,k}$, while \methodshortvar{} uploads $\C_{t,k}$ and $\nn_{t,k}$ 
\begin{table}[t]
	\centering
	\resizebox{\linewidth}{!}{
	\begin{tabular}{l >{\centering\arraybackslash}p{2.4cm} >{\centering\arraybackslash}p{1.8cm}}
		\toprule
		Method & Communication Overhead $\downarrow$ & Wallclock Time $\downarrow$ \\
		\midrule
		\multicolumn{3}{l}{\textbf{ResNet18 on CIFAR100 with 10 tasks, $M=5000$, $K_D=50$}} \\
        \midrule
		LANDER \cite{lander} & 79047.9 MB & \textgreater\; 10 h \;\;\;   \\
		\rowcolor{Thistle!20}	
		\methodshortvar{} (Ours) & 95.4 MB & \textless\; 3 min \\
		\rowcolor{Thistle!20}	
		\methodshort{} (Ours) & 955.6 MB & \textless\; 3 min \\
		\midrule
		\multicolumn{3}{l}{\textbf{ViT-B/16 on CIFAR100 with 10 tasks, $M=1250$, $K_D=10$}} \\
        \midrule
		PILoRA \cite{pilora} & 36.39 MB & \textgreater\; 20 h \;\;\;   \\
		\rowcolor{Thistle!20}	
		\methodshortvar{} (Ours) & 4.77 MB & \textless\; 10 min \\
		\rowcolor{Thistle!20}	
		\methodshort{} (Ours) & 60.1 MB & \textless\; 10 min \\
		\bottomrule
	\end{tabular}}
    \vspace{-10pt}
	\caption{Cost comparison with top-performing baselines after the training in the first stage.}
	\label{tab:cost_combined}
	\vspace{-10pt}
\end{table}
(assuming no dummy clients are used for simplicity). Feature statistics $\G_{t,k}$ and $\C_{t,k}$ are the aggregation of feature statistics from all local data samples. In real-world scenarios, each client $k$ typically has many data samples across various classes, making it difficult to reconstruct specific samples from the uploaded feature statistics. To support this, we conduct feature inversion experiments \cite{inversion2, ccvr} to validate this claim (\textbf{Appendix \ref{inversion_details}}). Label frequency $\nn_{t,k}$ is widely used in various FL methods \cite{ccvr, fedftg, fed_ensemble_d, fedgen, dfrd, fedcvae} and poses fewer privacy risks compared to data samples. Moreover, compatible privacy-preserving techniques, such as secure aggregation, can be integrated to further strengthen our method’s privacy protection, as the server requires only the aggregated Spatial Statistics rather than individual client values.

\section{Related Work}
\noindent
\textbf{Class-Incremental Learning.} Continual learning (CL) enables a learning system to acquire knowledge from sequentially arriving tasks without catastrophic forgetting \cite{cl_survey1}. In CL, Class-Incremental Learning (CIL) is the most popular setting due to its high relevance to real-world applications \cite{cl_survey1, mfcl}. In this setting, each new task introduces additional classes to the output space, progressively increasing the total number of classes. Existing CIL methods address catastrophic forgetting using various strategies, such as data replay \cite{cl_er2, cl_er3}, weight regularization \cite{cl_rt, cl_rt3}, knowledge distillation \cite{lwf, cl_kd1}, and network expansion \cite{dualprompt, coda, ease}. Although current CIL methods have made significant progress, they often struggle to perform well in the FCIL setting. This is due to the new challenges introduced by FL, such as data heterogeneity, communication constraints, and privacy concerns.

\noindent
\textbf{Federated Class-Incremental Learning.} GLFC \cite{fcl} is the first to introduce the FCIL setting and utilizes stored historical exemplars for local relation distillation. Various strategies are then proposed to make the best use of historical exemplars to mitigate catastrophic forgetting \cite{fcl_lga, fcl_replay1}. However, storing data from previous tasks is prohibited due to strict privacy regulations and policies in many privacy-sensitive scenarios, e.g., hospitals and medical research institutions \cite{privacy1, target, lander}. Therefore, the focus has shifted to more challenging exemplar-free scenarios. Existing state-of-the-art exemplar-free FCIL methods can be broadly classified into two categories: (1) Data-Free Knowledge Distillation (DFKD) approaches \cite{fedcil, mfcl, target, lander} and (2) Parameter-Efficient Fine-Tuning (PEFT) approaches \cite{fcl_prompt, fcl_prompt2, cprompt, pilora}. DFKD-based FCIL methods train a generator to generate synthetic samples based on the predictions of the global model. These synthetic samples are stored for training subsequent tasks. However, storing and transferring large volumes of synthetic data between the server and clients would incur significant storage and communication costs. PEFT-based FCIL methods store knowledge from previous stages in prompt \cite{fcl_prompt, fcl_prompt2, cprompt} or LoRA \cite{pilora} parameters and retain it using similarity matching or task arithmetic. However, these methods fail to fully exploit the generalization capabilities of pre-trained models \cite{revisiting} and remain vulnerable to spatial-temporal client drift, ultimately resulting in poor performance under high data heterogeneity. Besides, they suffer from computational inefficiency due to iterative backpropagation steps, placing a significant burden on client-side resources.

\vspace{-5pt}
\section{Conclusion and Future Work}
We propose a novel FCIL method, \methodshort{}, which unifies feature statistic aggregation across spatial and temporal dimensions. The aggregated statistics are immune to data heterogeneity and enable closed-form classifier updates at each stage. We also introduce \methodshortvar{}, a variant that achieves similar performance with lower communication costs. Extensive experiments show that our method outperforms state-of-the-art baselines, offering greater flexibility while reducing communication and computation overhead. In the future, we plan to extend \methodall{} to more FCL settings (e.g., asynchronous scenarios) and explore strategies for more efficient use of feature statistics.

\section*{Impact Statement}
This paper aims to advance the field of Federated Class-Incremental Learning (FCIL). FCIL has the potential to enable privacy-preserving continual learning in decentralized environments. It allows edge devices (e.g., smartphones, wearables, and IoT sensors) to collaboratively build powerful models that can incrementally learn new tasks or classes over time without sharing raw data. This strengthens user privacy and ensures compliance with data protection regulations (e.g., GDPR). Our research enhances the effectiveness and efficiency of FCIL, enabling broader applications.
\vspace{-10pt}
\begin{section}*{Acknowledgments}
This work is supported by the Chinese Academy of Sciences under grant No. XDB0690302, the National Key Research and Development Program of China (NO. 2024YFE0203200),  and the National Nature Science Foundation of China (NO. U24A20329).
\end{section}

\nocite{langley00}

\bibliography{example_paper}
\bibliographystyle{icml2024}

\newpage
\appendix
\onecolumn
\clearpage
\setcounter{page}{1}

\section{Proof of Proposition 3.1}
\label{proof}
\textbf{Proposition 3.1.}
\textit{For each task $\mathcal{T}_t$,}
\label{proposition 1}
\begin{align}
\hat{\boldsymbol{G}}_t = \sum_{i=1}^{c_t} \Bigg[\frac{n^{(i)}_t - 1}{K - 1}\sum_{k=1}^K \frac{\boldsymbol{C}_{t,k}^{(i)}\boldsymbol{C}_{t,k}^{(i)\T}}{n_{t,k}^{(i)}} - \frac{n^{(i)}_t - K}{n^{(i)}_t (K - 1)} \bigg( \sum_{k=1}^K \boldsymbol{C}_{t,k}^{(i)} \bigg) \bigg( \sum_{k=1}^K \boldsymbol{C}_{t,k}^{(i)} \bigg)^\top \Bigg],
\end{align}
\textit{is an unbiased plug-in estimator of the global gram matrix $\boldsymbol{G}_t$ when global feature set \text{~$\Omega=\{(\Phi(x),y) \mid(x,y)\in \cup_{k=1}^K \train_{t,k}\}$} is i.i.d. Here, $\boldsymbol{C}_{t,k}^{(i)}$ denotes the $i$-th column of $\boldsymbol{C}_{t,k}$ and $n_t^{(i)}$ denotes the total number of data samples from class $i$ in stage $t$, and $n_{t,k}^{(i)}$ denotes the number of samples from class $i$ for client $k$ at stage $t$.}
\begin{proof}
For simplicity, we omit the subscript $t$ in the proof. Under the i.i.d assumption of the global feature set ~$\{(\Phi(x),y) \mid(x,y)\in \cup_{k=1}^K D_{k}\}$, and letting $n_k^{(i)}$ denote the total number of data samples from class $i$ in client $k$, we have:
\begin{align}\label{formula_14}
     \mathbb{E} [\boldsymbol{C}_{k}^{(i)}]=\sum_{(x,i)\in D_{k}} \mathbb{E} [\Phi(x)]:=n_k^{(i)}\mu^{(i)}, \quad \mathbb{D} [\boldsymbol{C}_{k}^{(i)}]=\sum_{(x,i)\in D_{k}} \mathbb{D} [\Phi(x)]:=n_k^{(i)}\Sigma^{(i)}.
\end{align}
Here, $\mu^{(i)}$ and $\Sigma^{(i)}$ are the class-wise mean and covariance of the global feature set $\Omega$. Since $\boldsymbol{G} =\sum_{i=1}^c \boldsymbol{G}^{(i)} = \sum_{i=1}^c\sum_{k=1}^K\boldsymbol{G}^{(i)}_k$, we have $\mathbb{E}[\boldsymbol{G}]=\sum_{i=1}^c\mathbb{E}[\boldsymbol{G}^{(i)}]$ and
\begin{align}
\mathbb{E}[\boldsymbol{G}^{(i)}] &=\sum_{k=1}^K \mathbb{E}[\boldsymbol{G}^{(i)}_k] = \sum_{k=1}^K \mathbb{E}[\boldsymbol{X}^{(i)\T}_k \boldsymbol{X}^{(i)}_k] = \sum_{k=1}^K\sum_{(x,i)\in D_{k}}\mathbb{E} [\Phi(x)\Phi(x)^\T] \notag\\
&= \sum_{k=1}^K (n_k^{(i)}\mu^{(i)}\mu^{(i)\T} + n_k^{(i)}\Sigma^{(i)}) =n^{(i)}\mu^{(i)}\mu^{(i)\T}+n^{(i)}\Sigma^{(i)}\Big.
\end{align}
For clarity in the proof, we expand \(\mathbb{E}[\hat{\boldsymbol{G}}]\) and divide it into several parts as follows:
\begin{align}
    \mathbb{E}[\hat{\boldsymbol{G}}]=\sum_{i=1}^c\mathbb{E}[\hat{\boldsymbol{G}}^{(i)}]&=\sum_{i=1}^c\Big[\frac{n^{(i)}-1}{K-1}A_{1,i}-\frac{n^{(i)}-K}{n^{(i)}(K-1)}A_{2,i}\Big], \notag \\
    & \text{where} \quad A_{1,i}=\sum_{k=1}^K \frac{\mathbb{E}[\boldsymbol{C}_{k}^{(i)}\boldsymbol{C}_{k}^{(i)\T}]}{n_k^{(i)}}, A_{2,i} = \mathbb{E}\Big[\bigg( \sum_{k=1}^K \boldsymbol{C}_{k}^{(i)} \bigg) 
    \bigg( \sum_{k=1}^K \boldsymbol{C}_{k}^{(i)} \bigg)^\T\Big]. 
\end{align}
To proceed, we compute each term in above equation separately:
\begin{align}
    A_{1,i} &=\sum_{k=1}^K \frac{\mathbb{E}[\boldsymbol{C}_{k}^{(i)}\boldsymbol{C}_{k}^{(i)\T}]}{n_k^{(i)}} = \sum_{k=1}^K \frac{\mathbb{E} [\boldsymbol{C}_{k}^{(i)}]\mathbb{E}[\boldsymbol{C}_{k}^{(i)}]^\T + \mathbb{D}[\boldsymbol{C}_{k}^{(i)}]}{n_k^{(i)}} \notag \\
    & = \sum_{k=1}^K (n_k^{(i)}\mu^{(i)}\mu^{(i)\T} + \Sigma^{(i)}) = n^{(i)}\mu^{(i)}\mu^{(i)\T} + K\Sigma^{(i)} \\
     A_{2,i} &= \mathbb{E}[\bigg( \sum_{k=1}^K \boldsymbol{C}_{k}^{(i)} \bigg) 
    \bigg( \sum_{k=1}^K \boldsymbol{C}_{k}^{(i)} \bigg)^\T]=\mathbb{E}[\sum_{k=1}^K \boldsymbol{C}_{k}^{(i)}]\mathbb{E}[\sum_{k=1}^K \boldsymbol{C}_{k}^{(i)}]^\T+\mathbb{D}[\sum_{k=1}^K \boldsymbol{C}_{k}^{(i)}] \notag \\
    &=(\sum_{k=1}^K n_{k}^{(i)}\mu^{(i)})(\sum_{k=1}^K n_{k}^{(i)}\mu^{(i)})^\T+\sum_{k=1}^Kn_{k}^{(i)}\Sigma^{(i)}=(n^{(i)})^2\mu^{(i)}\mu^{(i)\T}+n^{(i)}\Sigma^{(i)}
\end{align}
Therefore, by substituting \( A_{1,i} \) and \( A_{2,i} \) back into the equation for \( \mathbb{E}[\hat{\boldsymbol{G}}] \), we get:
\begin{align}
    \hspace{-30pt}\mathbb{E}[\hat{\boldsymbol{G}}]&=\sum_{i=1}^c\Big [\frac{n^{(i)}-1}{K-1}(n^{(i)}\mu^{(i)}\mu^{(i)\T}\;\;+\; K\Sigma^{(i)})-\frac{n^{(i)}-K}{n^{(i)}(K-1)}((n^{(i)})^2\mu\mu^\T+n^{(i)}\Sigma^{(i)}) \Big]\notag \\
    &=\sum_{i=1}^c\Big [n^{(i)}\mu^{(i)}\mu^{(i)\T}+n^{(i)}\Sigma^{(i)}\Big ]=\sum_{i=1}^c\mathbb{E}[\boldsymbol{G}^{(i)}]=\mathbb{E}[\boldsymbol{G}]
\end{align}
\end{proof}

\section{i.i.d. assumption of global feature set $\Omega$}
\label{proof_iid}
In FCIL, \textbf{non-i.i.d. refers to label skew setting} where the label distributions $P_k(y)$ vary across clients, while the conditional distribution $P_k(x|y)$ remains the same for all clients \cite{target, lander}. Given the shared backbone $\Phi$, the conditional feature distribution $P_k(\Phi(x)|y)$ is also uniform across clients. Therefore, the global feature set $\Omega = \{(\Phi(x),y) \mid (x,y) \in \cup_{k=1}^K \mathcal{D}_{k} \}$ follows a mixture distribution: \vspace{-2pt}
\begin{align}
  P_{\Omega}(\Phi(X), Y) = \sum_{k=1}^K w_k P_k(\Phi(X), Y) = P(\Phi(X)|Y) \sum_{k=1}^K w_k P_k(Y) = P(\Phi(X)|Y) P_{avg}(Y),  
\end{align}
where $w_k$ is the proportion of data from client $k$ ($w_k=\frac{|D_k|}{|\cup_{k=1}^KD_k|}$), and $P_{avg}(Y)$ is the global average label distribution across all data.

The non-i.i.d. nature of the clients ($P_k(Y)$) only influences the shape of the final mixture distribution $P_{\Omega}$ (via $P_{avg}(Y)$), but it does not prevent samples drawn from $\Omega$ from being identically distributed according to $P_{\Omega}$. Therefore, the i.i.d. assumption of ${\Omega}$ in Proposition 3.1 (following the single mixture distribution $P_{\Omega}$) is valid within the label skew setting (non-i.i.d. in FCIL).

This explains why STSA-E shows strong robustness and comparable performance to STSA across varying data heterogeneity in our experiments.

\section{Estimation Guarantee of STSA-E}
\label{proof_stsa_e}
\textbf{Notation:} Let $\|\cdot\|_2$ denote the spectral norm, $\|\cdot\|_F$ represent the Frobenius norm, $\mathbb{E}$ indicate the expectation operator, $\mathbb{D}$ signify the variance operator, $\mathbf{0}$ be the zero vector, $\mathbf{I}_M$ denote the $M$-dimensional identity matrix, and $\operatorname{tr}(\cdot)$ represent the matrix trace operation.
We consider a noisy linear regression model and let $n$ represent the total number of collected samples until stage $t$, that is, $n=\sum_{i=1}^t n_i$ and $c_{\text{joint}}=\sum_{i=1}^{t}c_i$. In the subsequent discussion, we assume that, through sufficient continual learning and accumulation, the number of samples $n$ exceeds the feature dimension $M$, i.e., $n>M$. Specifically, we assume there exist ground-truth weight matrix $\boldsymbol{W}_t^0 \in \mathbb{R}^{M\times c_{\text{joint}}}$ and noise $\mathcal{E}_{joint} \in \mathbb{R}^{n\times c_{\text{joint}}}$ satisfying
\begin{align}
    \boldsymbol{Y}_{\text{~joint}}= \boldsymbol{X}_{\text{~joint}}\boldsymbol{W}^0_t+\mathcal{E}_{\text{~joint}}.
\end{align}
For each task $\mathcal{T}_t$, and for every $i \in\left\{1, \ldots, c_t\right\}$, we assume:

\textbf{Assumption 1 (Data Structure).} $\|\boldsymbol{X}_{t}\|_F\leq \sqrt{n_t} C_{\boldsymbol{X}}$, $\operatorname{tr}\left(\Sigma_t^{(i)}\right) \leq C_{\Sigma}$ and $\|\mu_t^{(i)}\|_2^2 \leq C_{\mu}$ for some constants $C_{\boldsymbol{X}},C_{\Sigma}, C_{\mu}>0$, where $\mu_t^{(i)}$ and $\Sigma_t^{(i)}$ are the class-wise mean and covariance of the global feature set $\Omega$ at stage $t$.

\textbf{Assumption 2 (Class Balance).} There exists a constant $0<C\ll c_t$ such that $c_t\sum_{i=1}^{c_t} n_t^{(i)2}\leq Cn_t^2$.

\textbf{Assumption 3 (Stability).} $\|\boldsymbol{W}_t^{0}\|_F \leq \sqrt{M}r$ for some constant $r>0$.

\textbf{Assumption 4 (Noise).} The $i$-th row of noise satisfies the following conditions: $\mathbb{E}(\mathcal{E}_{\text{~joint},i}^\top)=\mathbf{0}$, $\mathbb{E}\left(\mathcal{E}_{\text{~joint},i}^{\top} \mathcal{E}_{\text{~joint},i}\right)=\sigma^2 \boldsymbol{I}_{M}$ and $\mathcal{E}_{\text{~joint}}$ is independent of $\boldsymbol{X}_{\text{~joint}}$.

These four assumptions are mild in statistical learning theory and are consistent with (and often weaker than) those in related theoretical works \cite{ding2024understanding, peng2024icl}. We formalize these assumptions to ensure theoretical rigor while maintaining practical relevance. Here, we further explain the rationality of these assumptions:

\textbf{Assumptions 1 (Data Structure) and 3 (Stability)} are common boundedness conditions in theoretical analyses \cite{peng2024icl}. They simply ensure that feature and weight norms remain finite, which naturally holds in real-world scenarios. \textbf{Assumption 2 (Class Balance)} is a mild and flexible condition for global class distribution (considering all data across all clients): it does not require equal sample sizes across classes but only rules out extreme imbalance where one class dominates almost all samples (i.e., $\sum n_t^{(i) 2} \ll n_t^2$ ). This aligns with practical FCIL settings, where the global class distribution typically includes multiple classes with reasonable diversity. \textbf{Assumption 4 (Noise)} generalizes standard noise models in statistical learning. We only require zero-mean, homoscedastic, and feature-independent noise. This condition is less restrictive than many prior works that assume Gaussianity \cite{ding2024understanding, peng2024icl}, enabling broader applicability to real-world noise patterns without imposing stringent distributional constraints.

\textbf{Proposition 3.2.} 
\textit{Under Assumption 1-2, when considering $\hat{\boldsymbol{G}}_t$ as described in Proposition 3.1, we obtain the following results:
\begin{align}
    \mathbb{E}\left\|\hat{\boldsymbol{G}}_t-\boldsymbol{G}_t\right\|^2_F\leq C^{\prime}C_{\Sigma}^2 (\frac{K+1}{K-1})^2n_t^2 = O((\frac{K+1}{K-1})^2n_t^2),
\end{align}
where $C^{\prime}$ represents a constant that is indenpendent of ($n_t$, $K$).}

\begin{proof}
For simplicity, we omit the subscript $t$ in the proof.
\begin{align}
    \left\|\hat{\boldsymbol{G}}-\boldsymbol{G}\right\|^2_F & \leq c\sum_{i=1}^c\left\|\hat{\boldsymbol{G}}^{(i)}-\boldsymbol{G}^{(i)}\right\|^2_F \notag\\
    &= c\sum_{i=1}^{c}\left\|\frac{n^{(i)}-1}{K-1}\sum_{k=1}^K\frac{ \boldsymbol{C}_{k}^{(i)}\boldsymbol{C}_{k}^{(i)\T}}{n_{k}^{(i)}}-\frac{n^{(i)}-K}{n^{(i)}(K-1)}\bigg(\sum_{k=1}^K \boldsymbol{C}_{k}^{(i)}\bigg) 
    \bigg(\sum_{k=1}^K \boldsymbol{C}_{k}^{(i)}\bigg)^\T - \sum_{k=1}^K \boldsymbol{X}^{(i)\T}_{k}\boldsymbol{X}^{(i)}_{k}\right\|^2_F \notag\\
    &\leq (K+1)c\sum_{i=1}^c \left((\frac{n^{(i)}-K}{K-1})^2A^{(i)}_1+ (\frac{n^{(i)}-1}{K-1})^2\sum_{k=1}^K A^{(i)}_{2,k}\right),
\end{align}
where $A^{(i)}_1$ measures global random error, $A^{(i)}_{2,k}$ measures local random error,
\begin{align}
    A^{(i)}_1 = \left\|\boldsymbol{X}^{(i)\T}\boldsymbol{X}^{(i)}-\frac{1}{n^{(i)}}\bigg(\sum_{k=1}^K \boldsymbol{C}_{k}^{(i)}\bigg)\bigg(\sum_{k=1}^K \boldsymbol{C}_{k}^{(i)}\bigg)^\T\right\|^2_F, \quad A^{(i)}_{2,k} =  \left\|\boldsymbol{X}_k^{(i)\T}\boldsymbol{X}_k^{(i)}-\frac{1}{n_k^{(i)}}\boldsymbol{C}_{k}^{(i)}\boldsymbol{C}_{k}^{(i)\T}\right\|^2_F.
\end{align}
Noticing that the forms of these two terms are identical (except for the constants in front), we can use Lemma 3.5 to analyze them, which we assume that $\mathbb{E}[\boldsymbol{C}^{(i)}_k]:=n_k^{(i)}\mu^{(i)},  \mathbb{D}[\boldsymbol{C}^{(i)}_k]:=n_k^{(i)}\Sigma^{(i)}$. Then we have,
\begin{align}
    \mathbb{E}[A^{(i)}_1] \leq \frac{n^{(i)}-1}{n^{(i)}}\operatorname{tr}(\Sigma^{(i)})^2,\quad \mathbb{E}[A^{(i)}_{2,k}] \leq \frac{n^{(i)}_k-1}{n^{(i)}_k}\operatorname{tr}(\Sigma^{(i)})^2,
\end{align}
then we have 
\begin{align}
    \mathbb{E}\left\|\hat{\boldsymbol{G}}-\boldsymbol{G}\right\|^2_F&\leq C_{\Sigma}^2 (K+1)c\sum_{i=1}^c \left((\frac{n^{(i)}-K}{K-1})^2\frac{n^{(i)}-1}{n^{(i)}}+ (\frac{n^{(i)}-1}{K-1})^2\sum_{k=1}^K \frac{n^{(i)}_k-1}{n^{(i)}_k}\right) \notag\\
    & = C^{\prime\prime}C_{\Sigma}^2 \frac{(K+1)^2}{(K-1)^2} c\sum_{i=1}^c n^{(i)2} \leq C^{\prime}C_{\Sigma}^2 \frac{(K+1)^2}{(K-1)^2} n^2 = O((\frac{K+1}{K-1})^2n^2),
\end{align}
where we scale some constants such as $\frac{n^{(i)}}{n^{(i)}-1}$ as $C^{\prime\prime}$, which is a product factor of a finite number of quantities approximately equal to 1. Moreover, we utilize our class balanced condition to get another constant $C^{\prime}$. 
\end{proof}

\textbf{Remark.} Proposition 3.2 indicates that the error between $\hat{\boldsymbol{G}}_t$ and $\boldsymbol{G}_t$ is highly sensitive to small values of K, but the sensitivity diminishes as K increases.

After quantitatively analyzing the relationship between $\hat{\boldsymbol{G}}_t$ and $\boldsymbol{G}_t$, we can use Proposition 3.1 and 3.2 to further explore more profound theoretical properties between $\boldsymbol{W}_t =\left(\hat{\boldsymbol{G}}_{1: t}+\gamma \boldsymbol{I}\right)^{-1} \boldsymbol{C}_{1: t}$ and $
\boldsymbol{W}^{\star} =\left(\boldsymbol{G}_{1: t}+\gamma \boldsymbol{I}\right)^{-1} \boldsymbol{C}_{1: t}$.

\textbf{Theorem 3.3 (\methodshortvar{} plug-in error).}
\textit{Under Assumption 1-2, with high probability, we have
\begin{align}
    \frac{1}{n}\mathbb{E}\left\|\boldsymbol{X}_{\text{~joint}}\boldsymbol{W}_t-\boldsymbol{X}_{\text{~joint}}\boldsymbol{W}^{\star}\right\|_F\leq \frac{C^{\prime\prime}C_{\boldsymbol{X}}(C_{\Sigma})^{3/2}C_{\mu}^{1/2} \frac{K+1}{K-1}t}{n\left(\frac{\gamma^2}{n^2}+C^{\prime}\right)} = O\left(\frac{\frac{K+1}{K-1}t}{n(\frac{\gamma^2}{n^2}+1)}\right),
\end{align}
where $C^{\prime}, C^{\prime\prime}$ are the constants that are independent of $(n,t,K,\gamma)$.
}

\begin{proof}
Firstly, we expand the term in expectation:
\begin{align}
    \|\boldsymbol{X}_{\text{~joint}}\boldsymbol{W}_t&-\boldsymbol{X}_{\text{~joint}}\boldsymbol{W}^{\star}\|_F\leq \left\|\boldsymbol{X}_{\text{~joint}}\right\|_F\left\|\boldsymbol{W}_t-\boldsymbol{W}^{\star}\right\|_F \notag\\
    &= \left\|\boldsymbol{X}_{\text{~joint}}\right\|_F \|\left[(\hat{\boldsymbol{G}}_{1:t}+\gamma I)^{-1}-(\boldsymbol{G}_{1:t}+\gamma I)^{-1}\right]\boldsymbol{C}_{1:t}\|_F \notag\\
    &= \left\|\boldsymbol{X}_{\text{~joint}}\right\|_F \|(\hat{\boldsymbol{G}}_{1:t}+\gamma I)^{-1}(\boldsymbol{G}_{1:t}-\hat{\boldsymbol{G}}_{1:t})(\boldsymbol{G}_{1:t}+\gamma I)^{-1}\boldsymbol{C}_{1:t}\|_F \notag\\
    &\leq \left\|\boldsymbol{X}_{\text{~joint}}\right\|_F \|(\hat{\boldsymbol{G}}_{1:t}+\gamma I)^{-1}\|_2\|\boldsymbol{G}_{1:t}-\hat{\boldsymbol{G}}_{1:t}\|_F\|(\boldsymbol{G}_{1:t}+\gamma I)^{-1}\|_2\|\boldsymbol{C}_{1:t}\|_F \notag\\
    &\leq \frac{\|\boldsymbol{X}_{\text{~joint}}\|_F\|\boldsymbol{C}_{1:t}\|_F\|\boldsymbol{G}_{1:t}-\hat{\boldsymbol{G}}_{1:t}\|_F}{\gamma^2+C^{\prime}n^2} \leq \frac{C_{\boldsymbol{X}} \sqrt{tn}\|\boldsymbol{C}_{1:t}\|_F}{\gamma^2+C^{\prime}n^2}\sum_{i=1}^{t}\|\boldsymbol{G}_i-\hat{\boldsymbol{G}}_i\|_F,
\end{align}
where the first inequality is based on the property $\|A B\|_F \leq\|A\|_F\|B\|_F$, and the second inequality follows from $\|A B\|_F \leq\|A\|_2\|B\|_F$. The second equality holds because $A^{-1}-B^{-1}=A^{-1}(B-A) B^{-1}$. 

Regarding the third inequality, we consider $\|(\boldsymbol{G}_{1:t}+\gamma I)^{-1}\|_2$ firstly. Notice that
\begin{align}
    \|(\boldsymbol{G}_{1:t}+\gamma I)^{-1}\|_2 & = \frac{1}{\operatorname{min}_i{\lambda_i+\gamma}} \qquad \text{where}\ \lambda_i\ \text{is the $i$-th eigenvalue of}\ \boldsymbol{G}_{1:t},
\end{align}
Where $\boldsymbol{G}_{1: t}$ is, in fact, the sum of the outer products of $n$ vectors. According to Lemma 3.7, when the number of vectors $n$ exceeds the dimension $M$, with high probability, we can derive that $\min _i \lambda_i\geq O(n)$. The analysis result for $\left\|\left(\hat{\boldsymbol{G}}_{1: t}+\gamma \boldsymbol{I}\right)^{-1}\right\|_2$ is similar by Lemma 3.7-3.8. Therefore, we can postulate the existence of a certain positive constant $C^{\prime}$ such that 
\begin{align}
    \left\|\left(\boldsymbol{G}_{1: t}+\gamma \boldsymbol{I}\right)^{-1}\right\|_2 \cdot\left\|\left(\hat{\boldsymbol{G}}_{1: t}+\gamma \boldsymbol{I}\right)^{-1}\right\|_2 \leq \dfrac{1}{C^{\prime} n^2+\gamma^2}.
\end{align}

Regarding the fourth inequality, it is derived by considering the Cauchy inequality and the fact that $\left\|\boldsymbol{X}_{\text {joint }}\right\|_F \leq \sum_{i=1}^t\left\|\boldsymbol{X}_i\right\|_F$, where $\boldsymbol{X}_{\text {joint }}$ is the matrix obtained by concatenating the matrices $\boldsymbol{X}_i$.

By Cauchy inequality, we have 
\begin{align}
    \mathbb{E}[\|\boldsymbol{C}_{1:t}\|_F\sum_{i=1}^{t}\|\boldsymbol{G}_i-\hat{\boldsymbol{G}}_i\|_F] &=\sum_{i=1}^{t}\mathbb{E}[\|\boldsymbol{C}_{1:t}\|_F\|\boldsymbol{G}_i-\hat{\boldsymbol{G}}_i\|_F]\leq \sum_{i=1}^{t} \sqrt{\mathbb{E}[\|\boldsymbol{C}_{1:t}\|^2_F]\mathbb{E}[\|\boldsymbol{G}_i-\hat{\boldsymbol{G}}_i\|^2_F]}
\end{align}
Then, we bound $\mathbb{E}[\|\boldsymbol{C}_{1:t}\|^2_F]$,
\begin{align}
    \mathbb{E}\|\boldsymbol{C}_{1:t}\|_F^2 & \leq \mathbb{E}\|\sum_{i=1}^t\boldsymbol{C}_t\|_F^2 \leq \mathbb{E}[\sum_{i=1}^t\|\boldsymbol{C}_t\|_F]^2 \notag \leq t \sum_{i=1}^t\mathbb{E} [\|\boldsymbol{C}_t\|^2_F] \notag\\
    & = t\sum_{i=1}^t \mathbb{E}\operatorname{tr}(\boldsymbol{X}_i^\T\boldsymbol{Y}_i\boldsymbol{Y}_i^\T\boldsymbol{X}_i) = t\sum_{i=1}^t \mathbb{E}\operatorname{tr}(\boldsymbol{Y}_i\boldsymbol{Y}_i^\T\boldsymbol{X}_i\boldsymbol{X}_i^\T) \notag \\
    & = t\sum_{i=1}^t\sum_{j=1}^{c_i}n_i^{(j)}\left(\mu_i^{(j)\T}\mu_i^{(j)}+\operatorname{tr}\left(\Sigma^{(j)}\right)\right) \leq  C_{\Sigma}C_{\mu}tn.
\end{align}
By taking the result in Proposition 3.2, and combining the above parts, we have
\begin{align}
    \mathbb{E}\left\|\boldsymbol{X}_{\text{~joint}}\boldsymbol{W}_t-\boldsymbol{X}_{\text{~joint}}\boldsymbol{W}^{\star}\right\|_F \leq \frac{C_{\boldsymbol{X}}C^{1/2}_{\Sigma}C^{1/2}_{\mu} t}{(\frac{\gamma^2}{n}+C^{\prime}n)} \sum_{i=1}^{t}\sqrt{\mathbb{E}[\|\boldsymbol{G}_i-\hat{\boldsymbol{G}}_i\|^2_F]}\leq \frac{C^{\prime\prime}C_{\boldsymbol{X}}(C_{\Sigma})^{3/2}C_{\mu}^{1/2} \frac{K+1}{K-1}t}{\frac{\gamma^2}{n^2}+C^{\prime}},
\end{align}
where $C^{\prime\prime}$ is the constant in Proposition 3.2.
\end{proof}

\textbf{Remark.} Theorem 3.3 establishes that the plug-in error decreases as the average per-stage sample size $\frac{n}{t}$ increases.

\textbf{Corollary 3.4 (Estimation error).}
\textit{Under the Assumption 1-4, with high probability, we have
\begin{align}
    \frac{1}{n}\mathbb{E}\left\|\boldsymbol{X}_{\text{~joint}}\boldsymbol{W}_t-\boldsymbol{Y}_{\text{~joint}}\right\|_F \leq \frac{C^{\prime\prime}C_{\boldsymbol{X}}(C_{\Sigma})^{3/2}C_{\mu}^{1/2} \frac{K+1}{K-1}t}{n\left(\frac{\gamma^2}{n^2}+C^{\prime}\right)}+\frac{MRC_{\boldsymbol{X}}+\sigma}{\sqrt{n}},
\end{align}
where $C^{\prime}, C^{\prime\prime}$ are the constants that are independent of $(n,t,K,\gamma)$.
}

\begin{proof}
Consider the error decomposition:
\begin{align}
    \frac{1}{n}\mathbb{E}\left\|\boldsymbol{X}_{\text{~joint}}\boldsymbol{W}_t-\boldsymbol{Y}_{\text{~joint}}\right\|_F &\leq \frac{1}{n}\mathbb{E}\left\|\boldsymbol{X}_{\text{~joint}}\boldsymbol{W}_t-\boldsymbol{X}_{\text{~joint}}\boldsymbol{W}^{\star}\right\|_F+\frac{1}{n}\mathbb{E}\left\|\boldsymbol{X}_{\text{~joint}}\boldsymbol{W}^{\star}-\boldsymbol{Y}_{\text{~joint}}\right\|_F \notag\\
    &\leq \frac{1}{n}\mathbb{E}\left\|\boldsymbol{X}_{\text{~joint}}\boldsymbol{W}_t-\boldsymbol{X}_{\text{~joint}}\boldsymbol{W}^{\star}\right\|_F+\frac{1}{n}\sqrt{\mathbb{E}\left\|\boldsymbol{X}_{\text{~joint}}\boldsymbol{W}^{\star}-\boldsymbol{Y}_{\text{~joint}}\right\|^2_F},
\end{align}
where the first term has been analyzed in Theorem 3.3, and we introduce Lemma 3.6 to analyze the second term. Then
\begin{align}
    & \frac{1}{n}\mathbb{E}\left\|\boldsymbol{X}_{\text{~joint}}\boldsymbol{W}_t-\boldsymbol{X}_{\text{~joint}}\boldsymbol{W}^{\star}\right\|_F \leq \frac{C^{\prime\prime}C_{\boldsymbol{X}}(C_{\Sigma})^{3/2}C_{\mu}^{1/2} \frac{K+1}{K-1}t}{n\left(\frac{\gamma^2}{n^2}+C^{\prime}\right)} = O\left(\frac{\frac{K+1}{K-1}t}{n\left(\frac{\gamma^2}{n^2}+1\right)}\right).\\
    & \frac{1}{n}\sqrt{\mathbb{E}\left\|\boldsymbol{X}_{\text{~joint}}\boldsymbol{W}^{\star}-\boldsymbol{Y}_{\text{~joint}}\right\|^2_F} \leq \frac{MRC_{\boldsymbol{X}}+\sigma}{\sqrt{n}} = O\left(\frac{1}{\sqrt{n}}\right).
\end{align}
Combing this two parts, we can proof the Corollary 3.4.
\end{proof}

\textbf{Remark.} The most significant implication of Corollary 3.4 lies in the rate analysis: our \methodshortvar{} incurs an estimation efficiency loss of order $O\left(t n^{-1}\right)$, whereas the oracle estimator achieves the standard rate $O\left(n^{-1 / 2}\right)$. Crucially, when $t \ll \sqrt{n}$ (a mild scaling condition), the efficiency loss becomes asymptotically negligible. Thus, our method achieves comparable statistical performance to the oracle approach while requiring substantially lower communication overhead, establishing a favorable trade-off between precision and resource efficiency.

\textbf{Lemma 3.5.} 
\textit{Let $\mathbf{x}_1, \mathbf{x}_2, \cdots, \mathbf{x}_n$ be i.i.d. random vectors such that $\mathbf{x}_i \sim N(\mu, \Sigma)$ for $i=1,2, \cdots, n$, where $\Sigma=(\sigma_{ij})_{i,j=1}^d$ is a $d \times d$ positive covariance matrix. Define the matrix $\mathbf{M}=\sum_{i=1}^n \mathbf{x}_i \mathbf{x}_i^\T-\frac{1}{n}\left(\sum_{i=1}^n \mathbf{x}_i\right)\left(\sum_{i=1}^n \mathbf{x}_i\right)^\T$. Then, the expected value of the Frobenius norm of $\mathbf{M}$ satisfies the following equality:
$$
\mathbb{E}\left[\|\mathbf{M}\|_F^2\right] = \frac{n-1}{n} \operatorname{tr}(\Sigma)^2.
$$}
\begin{proof}
    Without loss of generality, we assume that $\mu=\mathbf{0}$. 
    
    By definition of Frobenius norm, we have $\mathbb{E}\left[\|\mathbf{M}\|^2_F\right]=\mathbb{E}\left[\operatorname{tr}\left(\mathbf{M}^2\right)\right]$, then we expand $\mathbf{M}^2$ obtain,
    \begin{align}
    \mathbf{M}^2 & =\left(\sum_{i=1}^n \mathbf{x}_i \mathbf{x}_i^\T-\frac{1}{n} \sum_{i=1}^n \sum_{j=1}^n \mathbf{x}_i \mathbf{x}_j^\T\right)\left(\sum_{k=1}^n \mathbf{x}_k \mathbf{x}_k^\T-\frac{1}{n} \sum_{k=1}^n \sum_{l=1}^n \mathbf{x}_k \mathbf{x}_l^\T\right) \notag\\
    & =\sum_{i=1}^n \sum_{k=1}^n\left(\mathbf{x}_i \mathbf{x}_i^\T\right)\left(\mathbf{x}_k \mathbf{x}_k^\T\right)-\frac{2}{n} \sum_{i=1}^n \sum_{k=1}^n \sum_{l=1}^n\left(\mathbf{x}_i \mathbf{x}_i^\T\right)\left(\mathbf{x}_k \mathbf{x}_l^\T\right)+\frac{1}{n^2} \sum_{i=1}^n \sum_{j=1}^n \sum_{k=1}^n \sum_{l=1}^n\left(\mathbf{x}_i \mathbf{x}_j^\T\right)\left(\mathbf{x}_k \mathbf{x}_l^\T\right).
    \end{align}
    (1). We compute $\mathbb{E}\left[\operatorname{tr}\left(\sum_{i=1}^n \sum_{k=1}^n\left(\mathbf{x}_i \mathbf{x}_i^\T\right)\left(\mathbf{x}_k \mathbf{x}_k^\T\right)\right)\right]$. Since $\mathbf{x}_i$ are independent and $\mathbb{E}\left[\mathbf{x}_i \mathbf{x}_j^\T\right]=\left\{\begin{array}{ll}\Sigma, & i=j \\ \mathbf{0}, & i \neq j\end{array}\right.$, we have:
    \begin{align}
        \mathbb{E}\left[\operatorname{tr}\left(\sum_{i=1}^n \sum_{k=1}^n\left(\mathbf{x}_i \mathbf{x}_i^\T\right)\left(\mathbf{x}_k \mathbf{x}_k^\T\right)\right)\right]=\sum_{i=1}^n \mathbb{E}\left[\operatorname{tr}\left(\left(\mathbf{x}_i \mathbf{x}_i^\T\right)^2\right)\right]+\sum_{i \neq k} \mathbb{E}\left[\operatorname{tr}\left(\left(\mathbf{x}_i \mathbf{x}_i^\T\right)\left(\mathbf{x}_k \mathbf{x}_k^\T\right)\right)\right],
    \end{align}
    then we calculate $\mathbb{E}\left[\operatorname{tr}\left(\left(\mathbf{x}_i \mathbf{x}_i^\T\right)^2\right)\right]$. Let $\mathbf{x}_i=\left(x_{i 1}, x_{i 2}, \cdots, x_{i d}\right)^\T$, $\left(\mathbf{x}_i \mathbf{x}_i^\T\right)_{j k}=x_{i j} x_{i k}$, and $\operatorname{tr}\left(\left(\mathbf{x}_i \mathbf{x}_i^\T\right)^2\right)=\sum_{j=1}^d \sum_{k=1}^d\left(x_{i j} x_{i k}\right)^2$. For a multivariate normal distribution $\mathbf{x} \sim N(\mathbf{0}, \Sigma)$, the fourth-moment formula is
    \begin{align}
        \mathbb{E}\left[x_j x_k x_u x_v\right]=\sigma_{j k} \sigma_{u v}+\sigma_{j u} \sigma_{k v}+\sigma_{j v} \sigma_{k u},
    \end{align}
    \begin{align}
        \mathbb{E}\left[\operatorname{tr}\left(\left(\mathbf{x}_i \mathbf{x}_i^\T\right)^2\right)\right] & =\sum_{j=1}^d \sum_{k=1}^d \mathbb{E}\left[\left(x_{i j} x_{i k}\right)^2\right] =\sum_{j=1}^d \sum_{k=1}^d\left(\sigma_{j k}^2+\sigma_{j j} \sigma_{k k}\right).
    \end{align}
    Then we calculate $\mathbb{E}\left[\operatorname{tr}\left(\left(\mathbf{x}_i \mathbf{x}_i^\T\right)\left(\mathbf{x}_k \mathbf{x}_k^\T\right)\right)\right]$ for $i \neq k$. Since $\mathbf{x}_i$ and $\mathbf{x}_k$ are independent, $\mathbb{E}\left[\left(\mathbf{x}_i \mathbf{x}_i^\T\right)\left(\mathbf{x}_k \mathbf{x}_k^\T\right)\right]=\mathbb{E}\left[\mathbf{x}_i \mathbf{x}_i^\T\right] \mathbb{E}\left[\mathbf{x}_k \mathbf{x}_k^\T\right]=\Sigma^2$. Then we have $\mathbb{E}\left[\operatorname{tr}\left(\left(\mathbf{x}_i \mathbf{x}_i^\T\right)\left(\mathbf{x}_k \mathbf{x}_k^\T\right)\right)\right]=\operatorname{tr}\left(\Sigma^2\right)=\sum_{j=1}^d \sum_{k=1}^d \sigma_{j k}^2$.
    \begin{align}
        \sum_{i=1}^n \mathbb{E}\left[\operatorname{tr}\left(\left(\mathbf{x}_i \mathbf{x}_i^\T\right)^2\right)\right] & =n \sum_{j=1}^d \sum_{k=1}^d\left(\sigma_{j k}^2+\sigma_{j j} \sigma_{k k}\right),\\
        \sum_{i \neq k} \mathbb{E}\left[\operatorname{tr}\left(\left(\mathbf{x}_i \mathbf{x}_i^\T\right)\left(\mathbf{x}_k \mathbf{x}_k^\T\right)\right)\right] & =n(n-1) \sum_{j=1}^d \sum_{k=1}^d \sigma_{j k}^2.
    \end{align}
    (2). We compute $E\left[\operatorname{tr}\left(\frac{2}{n} \sum_{i=1}^n \sum_{k=1}^n \sum_{l=1}^n\left(\mathbf{x}_i \mathbf{x}_i^\T\right)\left(\mathbf{x}_k \mathbf{x}_l^\T\right)\right)\right]$. We consider different cases based on the indices. When $k=l$, we have terms related to $\mathbb{E}\left[\left(\mathbf{x}_i \mathbf{x}_i^\T\right)\left(\mathbf{x}_k \mathbf{x}_k^\T\right)\right]$. When $k \neq l$, we use the independence of $\mathbf{x}_k$ and $\mathbf{x}_l$.
    \begin{align}
        \mathbb{E}\left[\operatorname{tr}\left(\frac{2}{n} \sum_{i=1}^n \sum_{k=1}^n \sum_{l=1}^n\left(\mathbf{x}_i \mathbf{x}_i^\T\right)\left(\mathbf{x}_k \mathbf{x}_l^\T\right)\right)\right]=\frac{2}{n}\left(n \sum_{j=1}^d \sum_{k=1}^d\left(\sigma_{j k}^2+\sigma_{j j} \sigma_{k k}\right)+2 n(n-1) \sum_{j=1}^d \sum_{k=1}^d \sigma_{j k}^2\right).
    \end{align}
    (3). We compute $E\left[\operatorname{tr}\left(\frac{1}{n^2} \sum_{i=1}^n \sum_{j=1}^n \sum_{k=1}^n \sum_{l=1}^n\left(\mathbf{x}_i \mathbf{x}_j^\T\right)\left(\mathbf{x}_k \mathbf{x}_l^\T\right)\right)\right]$. By considering all possible combinations of indices and using the independence of $\mathbf{x}_i$ and the properties of the normal distribution, we can show that
    \begin{align}
        \mathbb{E}\Bigg[\operatorname{tr}&\left(\frac{1}{n^2} \sum_{i=1}^n \sum_{j=1}^n \sum_{k=1}^n \sum_{l=1}^n\left(\mathbf{x}_i \mathbf{x}_j^\T\right)\left(\mathbf{x}_k \mathbf{x}_l^\T\right)\right)\Bigg] \notag\\
        &= \frac{1}{n^2}\left(n \sum_{j=1}^d \sum_{k=1}^d\left(\sigma_{j k}^2+\sigma_{j j} \sigma_{k k}\right)+4 n(n-1) \sum_{j=1}^d \sum_{k=1}^d \sigma_{j k}^2+\left(n^2-3 n+2\right) \sum_{j=1}^d \sum_{k=1}^d \sigma_{j k}^2\right).
    \end{align}
    After combining the above three parts, we have
    \begin{align}
        \mathbb{E}\left[\operatorname{tr}\left(\mathbf{M}^2\right)\right]=\frac{n-1}{n} \sum_{j=1}^d \sum_{k=1}^d \sigma_{j j} \sigma_{k k} = \frac{n-1}{n} \operatorname{tr}(\Sigma)^2.
    \end{align}
\end{proof}

\textbf{Lemma 3.6.} \textit{Suppose that $\boldsymbol{Y}=\boldsymbol{X} \boldsymbol{W}_0+\boldsymbol{\epsilon}$, where $\boldsymbol{Y} \in \mathbb{R}^{n \times c}, \boldsymbol{X} \in \mathbb{R}^{n \times M}$ is a random matrix whose rows $\boldsymbol{x}_i \sim \mathcal{N}(\mu, \Sigma)$ for $i=1, \cdots, n, \boldsymbol{W}_0 \in \mathbb{R}^{M \times c}$, and $\boldsymbol{\epsilon} \in \mathbb{R}^{n \times c}$ is a noise matrix with $\mathbb{E}(\boldsymbol{\epsilon})=\mathbf{0}$, $\mathbb{E}\left(\boldsymbol{\epsilon} \boldsymbol{\epsilon}^\T\right)=\sigma^2 \boldsymbol{I}_{n,}$ and $\boldsymbol{\epsilon}$ is independent of $\boldsymbol{X}$. Let $\boldsymbol{W}=\left(\boldsymbol{X}^\T \boldsymbol{X}+\gamma \boldsymbol{I}_M\right)^{-1} \boldsymbol{X}^\T \boldsymbol{Y}$ be the ridge estimator. Then
\begin{align}
    \mathbb{E}\|\boldsymbol{X} \boldsymbol{W}-\boldsymbol{Y}\|_F^2=\underbrace{\mathbb{E} \operatorname{tr}\left[\left(\boldsymbol{H}-\boldsymbol{I}_n\right)^2 \boldsymbol{X} \boldsymbol{W}_0 \boldsymbol{W}_0^\T \boldsymbol{X}^\T\right]}_{\text {Bias }}+\underbrace{\operatorname{tr}\left[\mathbb{E}\left(\boldsymbol{H}-\boldsymbol{I}_n\right)^2\right] \sigma^2}_{\text {Variance }} .
\end{align}  
where $\boldsymbol{H}=\boldsymbol{X}\left(\boldsymbol{X}^\T \boldsymbol{X}+\gamma \boldsymbol{I}_M\right)^{-1} \boldsymbol{X}^\T$. Besides, suppose further that: (1). The covariate matrix $\boldsymbol{X}$ satisfies $\|\boldsymbol{X}\|_F \leq \sqrt{n} C_{\boldsymbol{X}}$ for some constant $C_{\boldsymbol{X}}>0$ . (2). The true parameter matrix $\boldsymbol{W}_0$ satisfies $\left\|\boldsymbol{W}_0\right\|_F \leq \sqrt{M} R$ for some constant $R>0$. Then, we have:
\begin{align}
    \mathbb{E}\|\boldsymbol{X} \boldsymbol{W}-\boldsymbol{Y}\|_F^2 \leq n(M^2R^2C_{\boldsymbol{X}}^2+\sigma^2).
\end{align}
}

\begin{proof}
    First, substitute the expression of $\boldsymbol{W}$ into $\boldsymbol{X} \boldsymbol{W}-\boldsymbol{Y}$. Given $\boldsymbol{W}=\left(\boldsymbol{X}^\T \boldsymbol{X}+\gamma \boldsymbol{I}_M\right)^{-1} \boldsymbol{X}^\T \boldsymbol{Y}$ and $\boldsymbol{Y}=\boldsymbol{X} \boldsymbol{W}_0+\boldsymbol{\epsilon}$, we denote $\boldsymbol{H}=\boldsymbol{X}\left(\boldsymbol{X}^\T \boldsymbol{X}+\gamma \boldsymbol{I}_M\right)^{-1} \boldsymbol{X}^\T$, then we have
    \begin{align}
        \boldsymbol{X} \boldsymbol{W}-\boldsymbol{Y} & =\boldsymbol{X}\left(\boldsymbol{X}^\T \boldsymbol{X}+\gamma \boldsymbol{I}_M\right)^{-1} \boldsymbol{X}^\T \boldsymbol{Y}-\boldsymbol{Y}  =\left(\boldsymbol{H}-\boldsymbol{I}_n\right) \boldsymbol{Y} \notag\\
        &=\left(\boldsymbol{H}-\boldsymbol{I}_n\right)\left(\boldsymbol{X} \boldsymbol{W}_0+\boldsymbol{\epsilon}\right)  =\left(\boldsymbol{H}-\boldsymbol{I}_n\right) \boldsymbol{X} \boldsymbol{W}_0+\left(\boldsymbol{H}-\boldsymbol{I}_n\right) \boldsymbol{\epsilon}.
    \end{align}
    Consider $\mathbb{E}\|\boldsymbol{X} \boldsymbol{W}-\boldsymbol{Y}\|_F^2=\mathbb{E} \operatorname{tr}\left[(\boldsymbol{X} \boldsymbol{W}-\boldsymbol{Y})^\T(\boldsymbol{X} \boldsymbol{W}-\boldsymbol{Y})\right]$, since $\mathbb{E}(\boldsymbol{\epsilon})=\mathbf{0}$ and $\boldsymbol{\epsilon}$ is independent of $\boldsymbol{X}$, the cross-terms involving $\boldsymbol{\epsilon}$ and $\boldsymbol{X} \boldsymbol{W}_0$ will have zero expectation. Then
    \begin{align}
        \mathbb{E}\|\boldsymbol{X} \boldsymbol{W}-\boldsymbol{Y}\|_F^2 &= \mathbb{E} \operatorname{tr}\left\{\left[\left(\boldsymbol{H}-\boldsymbol{I}_n\right) \boldsymbol{X} \boldsymbol{W}_0\right]^\T\left[\left(\boldsymbol{H}-\boldsymbol{I}_n\right) \boldsymbol{X} \boldsymbol{W}_0\right]\right\} +\mathbb{E} \operatorname{tr}\left\{\left[\left(\boldsymbol{H}-\boldsymbol{I}_n\right) \boldsymbol{\epsilon}\right]^\T\left[\left(\boldsymbol{H}-\boldsymbol{I}_n\right) \boldsymbol{\epsilon}\right]\right\} \notag\\
        &=\underbrace{\mathbb{E} \operatorname{tr}\left[\left(\boldsymbol{H}-\boldsymbol{I}_n\right)^2 \boldsymbol{X} \boldsymbol{W}_0 \boldsymbol{W}_0^\T \boldsymbol{X}^\T\right]}_{\text {Bias }}+\underbrace{\operatorname{tr}\left[\mathbb{E}\left(\boldsymbol{H}-\boldsymbol{I}_n\right)^2\right] \sigma^2}_{\text {Variance }} .
    \end{align}
    \textbf{Bias Term.} Recall $\boldsymbol{H}=\boldsymbol{X}\left(\boldsymbol{X}^\T \boldsymbol{X}+\gamma \boldsymbol{I}_M\right)^{-1} \boldsymbol{X}^\T$. Using the identity: $\boldsymbol{H}-\boldsymbol{I}_n=\boldsymbol{X}\boldsymbol{X}^\T\left(\boldsymbol{X} \boldsymbol{X}^\T+\gamma \boldsymbol{I}_n\right)^{-1}-\boldsymbol{I}_n=-\gamma\left(\boldsymbol{X} \boldsymbol{X}^\T+\gamma \boldsymbol{I}_n\right)^{-1}$. We have $\left(\boldsymbol{H}-\boldsymbol{I}_n\right)^2=\gamma^2\left(\boldsymbol{X} \boldsymbol{X}^\T+\gamma \boldsymbol{I}_n\right)^{-2}$. The bias term becomes:
    \begin{align}
        \mathbb{E} \operatorname{tr}\left[\gamma^2\left(\boldsymbol{X} \boldsymbol{X}^\T+\gamma \boldsymbol{I}_n\right)^{-2} \boldsymbol{X} \boldsymbol{W}_0 \boldsymbol{W}_0^\T \boldsymbol{X}^\T\right]&=\gamma^2 \mathbb{E}\left[\operatorname{tr}\left(\boldsymbol{W}_0^\T \boldsymbol{X}^\T\left(\boldsymbol{X} \boldsymbol{X}^\T+\gamma \boldsymbol{I}_n\right)^{-2} \boldsymbol{X} \boldsymbol{W}_0\right)\right] \notag\\
        & \leq \gamma^2 \mathbb{E}\left[\operatorname{tr}\left(\boldsymbol{X}^\T\left(\boldsymbol{X} \boldsymbol{X}^\T+\gamma \boldsymbol{I}_n\right)^{-2} \boldsymbol{X} \right)\|\boldsymbol{W}_0\|_F^2\right] \notag \\
        & \leq \gamma^2 MR^2\mathbb{E}\left[\operatorname{tr}\left(\boldsymbol{X}^\T\left(\boldsymbol{X} \boldsymbol{X}^\T+\gamma \boldsymbol{I}_n\right)^{-2} \boldsymbol{X} \right)\right] \notag
    \end{align}
    Using properties of semi-positive matrices and the identity:
    \begin{align}
        \mathbb{E}\left[\operatorname{tr}\left(\boldsymbol{X}^\T\left(\boldsymbol{X} \boldsymbol{X}^\T+\gamma \boldsymbol{I}_n\right)^{-2} \boldsymbol{X} \right)\right] \leq \frac{n M C_{\boldsymbol{X}}^2}{\gamma^2}
    \end{align}
    Then, the bias term is bounded by:
    \begin{align}
        \text{~Bias} \leq \gamma^2 M^2R^2\operatorname{tr}\left(\mathbb{E}\left[\boldsymbol{X}^\T\left(\boldsymbol{X} \boldsymbol{X}^\T+\gamma \boldsymbol{I}_n\right)^{-2} \boldsymbol{X} \right]\right)\leq n M^2 R^2 C_{\boldsymbol{X}}^2
    \end{align}
    \textbf{Variance term.} Using the identity: $\boldsymbol{H}-\boldsymbol{I}_n=-\gamma\left(\boldsymbol{X} \boldsymbol{X}^\T+\gamma \boldsymbol{I}_n\right)^{-1}$, we have
    \begin{align}
        \operatorname{tr}\left[\mathbb{E}\left(\boldsymbol{H}-\boldsymbol{I}_n\right)^2\right] \sigma^2=\gamma^2 \sigma^2 \mathbb{E}\left[\operatorname{tr}\left(\left(\boldsymbol{X} \boldsymbol{X}^\T+\gamma \boldsymbol{I}_n\right)^{-2}\right)\right].
    \end{align}
    For semi-positive matrix $\boldsymbol{X} \boldsymbol{X}^\T$, we have $\operatorname{tr}\left(\mathbb{E}\left[\left(\boldsymbol{X} \boldsymbol{X}^\T+\gamma \boldsymbol{I}_n\right)^{-2}\right]\right) \leq \dfrac{n}{\gamma^2}$.
    Thus, the variance term is bounded by:
    \begin{align}
        \text{~Variance} \leq n\sigma^2.
    \end{align}
    Combining the bias and variance bounds, we will get the final bound.
\end{proof}

\textbf{Remark.} The result obtained in Lemma 3.6 aligns with the works of \citet{hsu2012random} and \citet{mourtada2022elementary}. For clarity of exposition and to stay within the theoretical framework of \citet{peng2024icl}, we have rederived a rough bound.

\textbf{Lemma 3.7(\citet{wainwright2019high}, Theorem 6.1).} \textit{Let $\mathbf{X} \in \mathbb{R}^{n \times d}$ be drawn according to the $\Sigma$-Gaussian ensemble. Then for all $\delta>0$, the maximum singular value $\sigma_{\max}(\mathbf{X})$ satisfies the upper deviation inequality,
\begin{align}
    \mathbb{P}\left[\frac{\sigma_{\max }(\mathbf{X})}{\sqrt{n}} \geq \lambda_{\max }(\sqrt{\Sigma})(1+\delta)+\sqrt{\frac{\operatorname{tr}(\Sigma)}{n}}\right] \leq e^{-n \delta^2 / 2}.
\end{align}
Moreover, for $n \geq d$, the minimum singular value $\sigma_{\min }(\mathbf{X})$ satisfies the analogous lower deviation inequality,
\begin{align}
    \mathbb{P}\left[\frac{\sigma_{\min }(\mathbf{X})}{\sqrt{n}} \leq \lambda_{\min }(\sqrt{\Sigma})(1-\delta)-\sqrt{\frac{\operatorname{tr}(\Sigma)}{n}}\right] \leq e^{-n \delta^2 / 2},
\end{align}
where $\lambda_{\max}(\cdot), \lambda_{\min}(\cdot)$ means the maximum and minimum eigenvalue respectively.}

\textbf{Remark.} Lemma 3.7 demonstrates that the following statement holds, which is used in the proof of Theorem 3.3. With a probability of at least $1-e^{-n \delta^2 / 2}$, 
\begin{align}
    \lambda_{\min}(\mathbf{X}\mathbf{X}^{\T}) \geq n\left(\lambda_{\min }(\sqrt{\Sigma})(1-\delta)-\sqrt{\frac{\operatorname{tr}(\Sigma)}{n}}\right)^2 = O(n).
\end{align}

\textbf{Lemma 3.8(\citet{wainwright2019high}, Exercise 6.1).} \textit{Given two symmetric matrices $\mathbf{A}$ and $\mathbf{B}$, we have
\begin{align}
    \left|\lambda_{\max }(\mathbf{A})-\lambda_{\max }(\mathbf{B})\right| \leq\|\mathbf{A}-\mathbf{B}\|_2 \quad \text { and } \quad\left|\lambda_{\min }(\mathbf{A})-\lambda_{\min }(\mathbf{B})\right| \leq\|\mathbf{A}-\mathbf{B}\|_2.
\end{align}}
\textbf{Remark.} Lemma 3.8 demonstrates that the following statement, which is used in the proof of Theorem 3.3. 
\begin{align}
    \left|\lambda_{\min }(\mathbf{A})\right|&=\left|\lambda_{\min }(\mathbf{B})-(\lambda_{\min }(\mathbf{B})-\lambda_{\min }(\mathbf{A}))\right| \geq \left|\lambda_{\min }(\mathbf{B})\right| - \left|\lambda_{\min }(\mathbf{B})-\lambda_{\min }(\mathbf{A})\right| \geq \left|\lambda_{\min }(\mathbf{B})\right| - \|\mathbf{A}-\mathbf{B}\|_2.
\end{align}

\section{Pseudo-Code}
We provide the the summary of the entire \methodall{} procedure with pseudocode in Algorithm \ref{alg:USTA}.
\SetKwInput{kwInput}{Input}
\SetKwInput{kwOutput}{Output}

\begin{algorithm}[t]
\LinesNumberedHidden 
\caption{Our \methodall{} for FCIL}
\label{alg:USTA}
\kwInput{Datasets: $\left\{\train_{1:T,1:K}\right\}$,  initial model: $\theta$, initial random seed: \texttt{SEED}.}
\textbf{Initialization}: server distributes the initial model and the initial random seed \texttt{SEED} to all clients. Each client initializes a random mapping layer $R$ using \texttt{SEED}. 

\ForEach{each task $t = 1,\cdots,T$}{
    \If{$t = 1$}{
                Update $\theta$ in federated manner. Fix the feature extractor $F$. 
    }
    \ForEach{\text{each client} $i = 1,\cdots,K$}{
        \If{\methodshortvar{}}{
            Extract Spatial Statistics $\C_{t,k}$ and $\nn_{t,k}$;
        }
        \Else{
        Extract Spatial Statistics $\{\G_{t,k},\C_{t,k}\}$;
        }
    }
    Spatial Statistics Aggregation via Eq. \ref{spatial_aggregation}; \\
    Temporal Statistics Aggregation via Eq. \ref{temporal_aggregation}; \\
    \If{\methodshortvar{}}{
        \hspace{4pt}Obtain $\hat{\G}_t$ via Eq. \ref{estimated_G}; \\
        Update classifier $\W_t$ via Eq. \ref{update_classifier2};
    }
    \Else{
        Update classifier $\W_t$ via Eq. \ref{update_classifier};
        }
}
\textbf{return} $\theta_t=\{F,R,\boldsymbol{W}_t\}$
\end{algorithm}

\section{Experimental Details}
\label{exp_details}
\subsection{Evaluation Metrics}
\label{appendix:metrics}
In this work, we employ three widely-used metrics in CIL: average incremental accuracy ($A_{\text{avg}}$), final average accuracy ($A_T$), and average forgetting ($F_T$). $A_{\text{avg}}$ denotes the average performance across all incremental stages, $A_T$ is the accuracy averaged over all $T$ tasks after completing task $\mathcal{T}_T$, and $F_T$ represents the average performance decline on previous tasks over all $T$ tasks. Let $A_{t,\tau}$ represent the classification accuracy on the $\tau$-th task after training on the $t$-th task. After training on the $t$-th task, we calculate the evaluation metrics as follows:
\begin{align}
    A_{\text{avg}} &=\sum_{t=1}^T\frac{1}{t}\sum_{\tau=1}^tA_{t,\tau}, \\
    A_T &= \frac{1}{T}\sum_{\tau=1}^TA_{T,\tau}, \\
    F_T &= \frac{1}{T-1}\sum_{\tau=1}^{T-1}\max_{\tau^{\prime}\in\{1,\cdots,T-1\}}A_{\tau^{\prime},\tau}-A_{T,\tau}.
\end{align}

\subsection{Client Training Details}
\label{appendix:client_training}
For the training-from-scratch setting, we follow the configurations from \citet{lander}, using ResNet18 \cite{resnet} as the backbone. Each client is trained using a batch size of 128 for 100 communication rounds, with 2 local training epochs per round. For CIFAR100 \cite{c10}, we employ the SGD optimizer with a learning rate of 0.04, a momentum of 0.9, and a weight decay of \(5 \times 10^{-4}\). For Tiny-ImageNet \cite{tin}, we use a learning rate of 0.1, a weight decay of \(2 \times 10^{-4}\), and a multi-step learning rate scheduler that reduces the learning rate by a factor of 10 at the 50th and 75th communication rounds.

When training from pre-trained model, we incorporate an adapter into each layer of pre-trained ViT via residual connections to facilitate parameter-efficient tuning \cite{revisiting}. Specifically, an adapter typically consists of a downsampled MLP layer $W_{down}\in \mathbb{R}^{d\times r}$, a non-linear activation function ReLU, and an upsampled MLP layer $W_{up}\in \mathbb{R}^{d\times r}$. We formulate the output of the MLP when using the adapter as follows:
\begin{equation}
\mathbf{x}_o=\mathrm{MLP}(\mathbf{x}_i)+s * \mathrm{ReLU}(\mathbf{x}_iW_{down})W_{up},
\end{equation}
where $\mathbf{x}_i$ represent the input, $\mathbf{x}_o$ denote the output, and $s$ be the scale factor.
By default, the scale factor \(s\) is set to 0.1, and the projection dimension \(r\) for the adapter is fixed at 64. For clients' local training, we use a learning rate of 0.2 and a weight decay of \(5 \times 10^{-4}\). The model is trained with a batch size of 128 for 2 local epochs. It runs for 10 communication rounds on CIFAR100 \cite{c10} and 20 communication rounds on ImageNet-R \cite{imagenetr}.

\subsection{Additional Results}
\label{full_ablation}
\begin{figure}[h]
\centering
\includegraphics[width=1\linewidth]{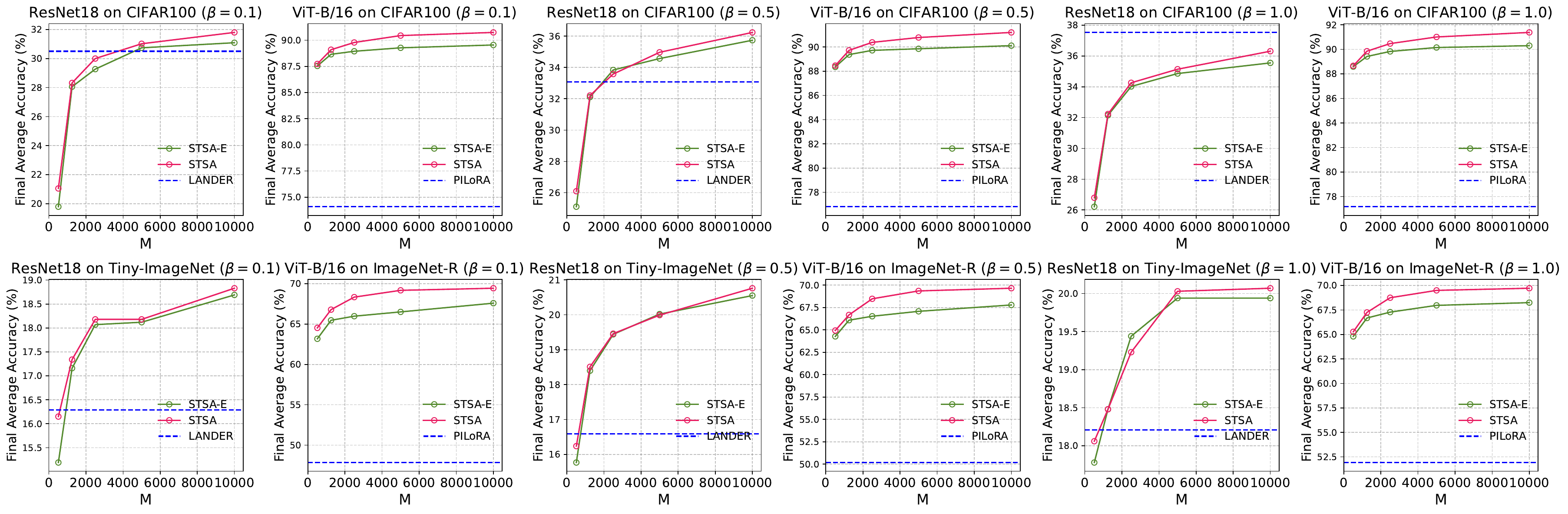}
\caption{
Ablation study on the dimension of random features ($M$). For ResNet18, $M$ varies across \{512, 1250, 2500, 5000, 10000\}, while for ViT, it varies across \{768, 1250, 2500, 5000, 10000\}. $M = 512$ corresponds to the raw feature dimension of ResNet18, and $M = 768$ corresponds to the raw feature dimension of ViT, indicating no random mapping is applied. The performance of the top-performing baselines (LANDER and PILoRA) is marked with a blue horizontal line in the figure.
}
\label{fig:all_M}
\end{figure}

\begin{figure}[h]
 \centering
\includegraphics[width=1\linewidth]{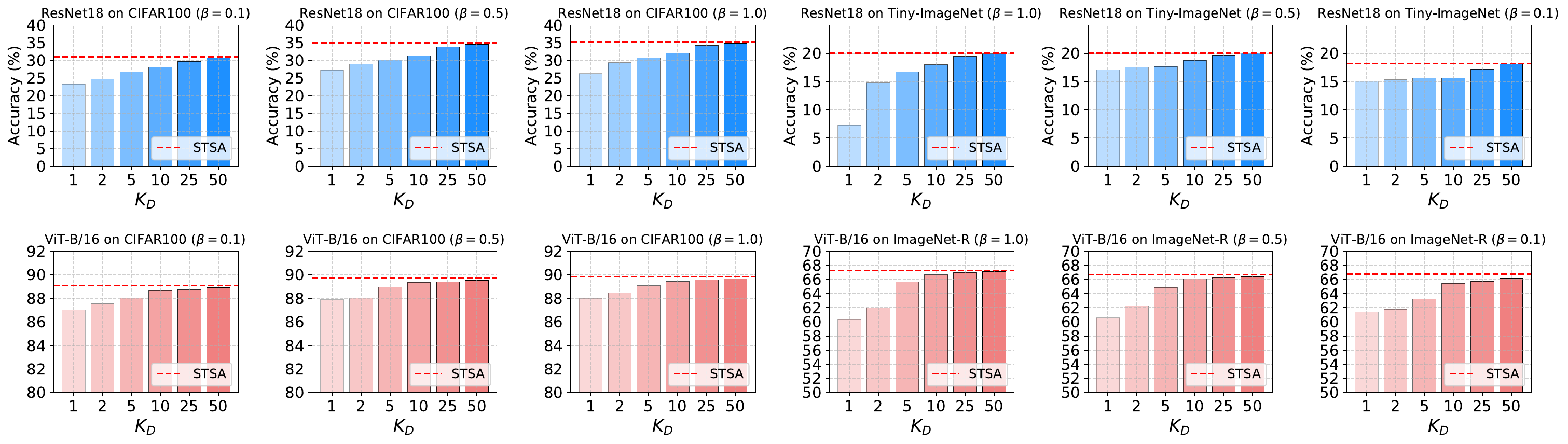}

\caption{Ablation study on the number of dummy clients ($K_D$). We test $K_D$ values in $\{1, 2, 5, 10, 25, 50\}$. When $K_D = 1$, no dummy clients are used. The performance of \methodshort{} is shown as a red horizontal line in the figure. Dummy clients are introduced to address scenarios with insufficient client numbers (e.g., cross-soil settings). When the number of clients is already large (e.g, cross-device settings), dummy clients are not needed. }
\label{fig:all_KD}
\end{figure}

Full results of ablations study are shown in Figure \ref{fig:all_M} and Figure \ref{fig:all_KD}.

\section{Privacy Discussion}
\label{privacy_detail}
In our \methodall{}, uploading Spatial Statistics may pose privacy concerns. Specifically, \methodshort{} uploads $\G_{t,k}$ and $\C_{t,k}$, while \methodshortvar{} uploads $\C_{t,k}$ and $\nn_{t,k}$ (assuming no dummy clients are used for simplicity). Feature statistics $\G_{t,k}$ and $\C_{t,k}$ are the aggregation of feature statistics from all local data samples. In real-world scenarios, each client $k$ typically has many data samples across various classes, making it difficult to reconstruct specific samples from the uploaded feature statistics. To support this, we conduct feature inversion experiments \cite{inversion2, ccvr} to validate this claim. Label frequency $\nn_{t,k}$ has been widely used in numerous FL methods \cite{ccvr, fedftg, fed_ensemble_d, fedgen, dfrd, fedcvae}, and label frequency presents fewer privacy risks compared to data samples. 


\subsection{Feature Inversion Details}
\label{inversion_details}
\begin{figure*}[t]
 \centering  
 \resizebox{\linewidth}{!}{  
\includegraphics[width=0.7\linewidth]{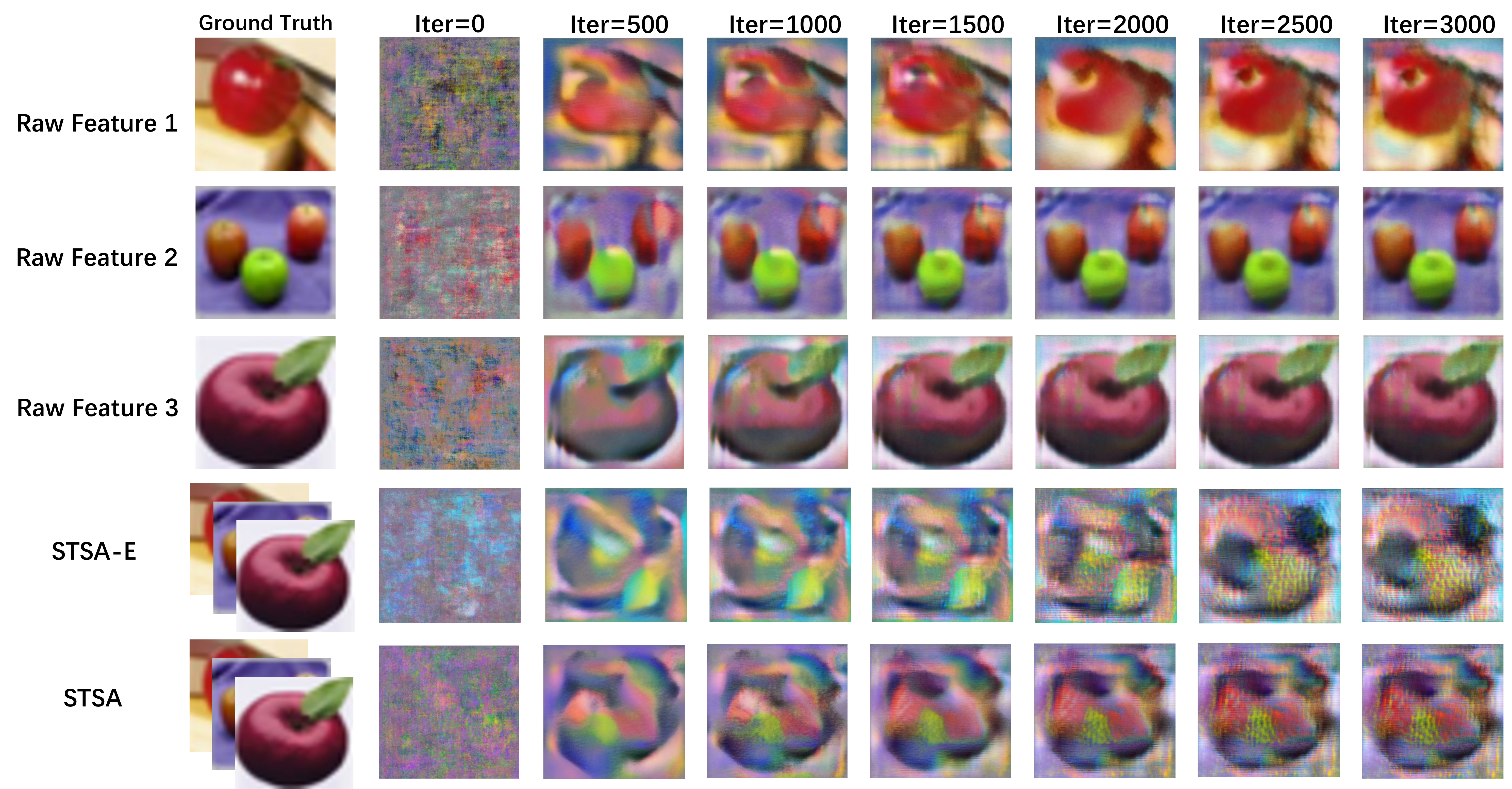}}  
\caption{  
Results of feature inversion on CIFAR100. Assume a client has only three "apple" samples. An attacker tries to reconstruct a data sample from this client. When the attacker accesses the \textbf{Raw Features} of a specific sample, the reconstructed results are clear. However, using our uploaded statistics, the reconstruction quality significantly deteriorates.}  
\label{fig:inversion}  
\end{figure*}

Here, we consider the scenario where the target client $k$ has only a few samples (3 samples) from the same class ("apple" in CIFAR100), and the attacker utilizes the uploaded Spatial Statistics to attempt reconstruction of a specific training data sample using feature inversion technology, as shown in \citet{inversion2, ccvr}. In this case, the uploaded Spatial Statistics is a mixture of features from a few samples (3 samples) within the same class, making it less mixed and thus more advantageous for the attacker. In addition to using the uploaded Spatial Statistics for feature inversion, we also provide experimental results for inversion using raw representations, where each raw feature corresponds to a single input sample. Table \ref{inversion_tab} presents the quantitative similarity metrics between the ground truth and the reconstructed results. Higher PSNR and SSIM values and lower LPIPS values indicate greater similarity to the ground truth, demonstrating better attack performance. For our \methodall{}, we compute similarity measures between the inversion result and each real sample, then average them to derive the final results. Figure \ref{fig:inversion} presents the visualization results of feature inversion. When using the raw feature of a specific data sample, the inversion produces clear visual outputs with higher quantitative similarity metrics. In contrast, using our uploaded Spatial Statistics for feature inversion leads to significantly worse metrics. The visual results consist of irregular color patches, making it hard to discern information related to the original data samples.

\subsection{Compatibility with Privacy-Preserving Techniques}
\label{privacy_tech}
Secure aggregation is especially suitable for our \methodall{}, since the server only needs the aggregated Spatial Statistics, not the individual value from each client. Additionally, we can add random noise to the uploaded Spatial Statistics to enhance the privacy-preserving capabilities of our method \cite{fedpretrained, dbe}. 

\begin{table}[h]
\vspace{-10pt}
\caption{Quantitative similarity measures of inversion results. For our \methodall{}, we compute similarity measures between the inversion result and 3 real samples, then average them to obtain the final results.}
\vspace{8pt}
\centering
\resizebox{0.5\linewidth}{!}{
\begin{tabular}{cc|c|c}
\toprule
\multirow{2}{*}{Reconstruction Source}  & \multicolumn{3}{c}{Similarity Measure} \\
\cmidrule(lr){2-4}
& $PSNR \downarrow$ & $SSIM \downarrow$ & $LPIPS \uparrow$\\
\midrule
Raw Feature 1 & 18.43 & 0.68 & 0.26  \\
Raw Feature 2 & 22.15 & 0.77 & 0.14  \\
Raw Feature 3 & 19.84 & 0.79 & 0.19  \\
\rowcolor{Thistle!20}	
\methodshortvar{}          & \underline{11.43} & \underline{0.34} & \underline{0.79}  \\
\rowcolor{Thistle!20}	
\methodshort{}          & \textbf{11.18} & \textbf{0.26} & \textbf{0.84}  \\
\bottomrule
\end{tabular}
}
\vspace{-10pt}
\vspace{-10pt}
\label{inversion_tab}
\end{table}

\vspace{-10pt}  

\end{document}